\documentclass[10pt,twocolumn,letterpaper]{article}

\usepackage[pagenumbers]{cvpr}

\usepackage{placeins}
\usepackage{algorithm}
\usepackage{algpseudocode}
\usepackage{subcaption}

\usepackage{adjustbox}

\usepackage[table]{xcolor}  %
\usepackage{booktabs}       %
\usepackage{makecell}       %

\usepackage{tabularx}       %
\usepackage{pifont}         %

\usepackage{pgfplots}
\pgfplotsset{compat=1.18}

\DeclareRobustCommand{\req}[1]{%
  \tikz[baseline=(char.base)]{\node[shape=circle,draw,inner sep=0.8pt]
  (char) {\scriptsize #1};}}

\newcommand{\cmark}{\checkmark}
\newcommand{\xmark}{$\times$}

\newcommand{\sees}[1]{\multicolumn{6}{@{}>{\columncolor{gray!10}[0pt][0pt]}l@{}}{\quad\textcolor{black!60}{\itshape #1}}}
\newcommand{\corner}[1]{{\scriptsize\textcolor{black!60}{\itshape #1}}}

\newcommand{\method}{DesignOrbit}

\newcommand{\archvis}{ArchVis}

\usepackage{microtype}

\definecolor{cvprblue}{rgb}{0.21,0.49,0.74}
\usepackage[pagebackref,breaklinks,colorlinks,allcolors=cvprblue]{hyperref}

\title{\method: \\Exploring the Unseen Sides of a Design in Context from a Single Image}

\author{
\begin{tabular}{ccc}
I-Ting Tsai & Abe Davis & Bharath Hariharan \\
Cornell University & Cornell University & Cornell University \\
{\tt\small itingtsai@cs.cornell.edu} &
{\tt\small abedavis@cornell.edu} &
{\tt\small bharathh@cs.cornell.edu}
\end{tabular}
\\[0.5em]
{\small Project Page:
\href{https://itingtsai.github.io/DesignOrbit/}
{\texttt{https://itingtsai.github.io/DesignOrbit/}}}
}

\begin{document}

\twocolumn[{%
\renewcommand\twocolumn[1][]{#1}%
\maketitle
\vspace{-3em}
\begin{center}
  \centering
  \captionsetup{type=figure}
  \includegraphics[width=0.95\textwidth]{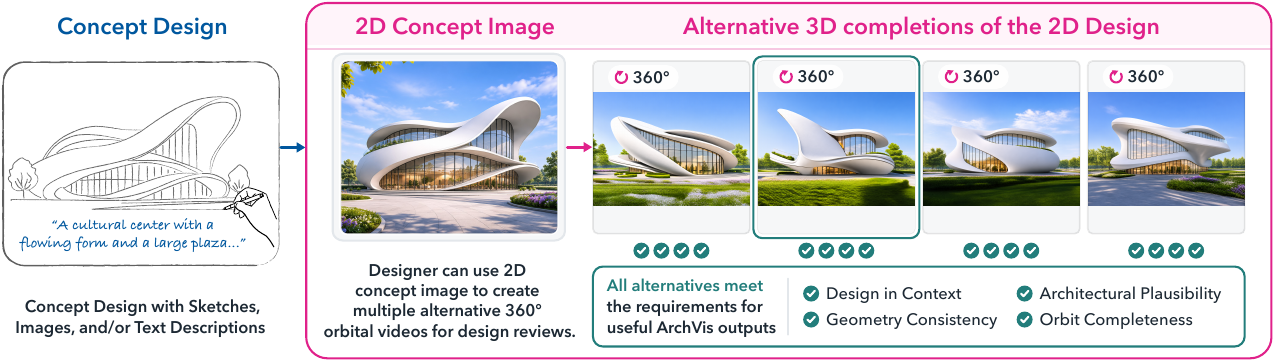}
  \captionof{figure}{
    \textbf{\method{} generates alternative 360$^\circ$ orbits from a single concept image.}
    Each orbit proposes a different 3D completion of the design while addressing four ArchVis requirements (Sec.~\ref{sec:background}): context, architectural plausibility, geometric consistency, and orbit completeness. Designers can compare alternatives and select one for detailed 3D modeling (outlined).
  }
  \label{fig:workflow}
\end{center}%
}]

\raggedbottom

\begin{abstract}
Rapid prototyping and pre-visualization are critical in architectural design. Generative models have potential to help with these tasks by letting architects turn simple image and text prompts into pre-visualizations of a design concept rendered in the context of its intended environment from different views. However, the objectives of architectural visualization (\archvis{}) are more specific than those used to train video models. As a result, current video models fall short on multiple fronts for \archvis{} applications in practice. In this paper, we start by identifying requirements for useful \archvis{} outputs, then develop a novel training pipeline based on those requirements, which we use to train \emph{DesignOrbit}, a model for generating $360^\circ$ orbital videos of building rendered in-context with an intended environment based on inputs that are well-suited for the iterative design processes used by real architects. 
DesignOrbit generates videos revealing unseen sides with consistent
geometry, completing $360^\circ$ orbits over $3$ times as often as
the strongest baseline
in SfM-based evaluations of the identified geometric requirements, and in a user study with real architects we find a preference for DesignOrbit over baseline models in $>90\%$ of comparisons.

\end{abstract}

\section{Introduction}
\label{sec:intro}

When designing buildings, pavilions, and public art installations, early
designs often exist as concept sketches or single images. Designers use
these representations to communicate ideas with their clients and review
the form and spatial relationships of a proposal~\cite{AIA}.
However, a single image provides limited information: a concept must
also be considered from its side and back views and in relation to its
surroundings, and the same front view admits many possible rear designs.

Architectural visualization (ArchVis) helps designers explore and
communicate proposals throughout the design process. At the conceptual
and schematic stages we focus on, generative images can support this
exploration before detailed 3D models are available
~\cite{lee2024generative,paananen2024ideation}.
A single image, however, cannot show how the proposed design looks
from other viewpoints or how it relates to its surroundings from
those views. Animated visualizations can reveal these aspects by
moving the camera around the design. Producing such animations through
a conventional 3D workflow requires modeling the design, assigning
materials, constructing its surroundings, setting up lighting, and
rendering~\cite{borkowski2023rendering}.
Building and updating this 3D scene takes time and effort, making
animated views costly to produce for early concepts that are continuously being iterated upon.

Given this motivation and recent advances in video generation, a natural goal is to turn a single
image and a text description of a design and its setting into a $360^\circ$ orbital video (\cref{fig:workflow}). 
However, the goals of ArchVis go beyond just realistic video generation.
Our key insight is that
useful ArchVis outputs require several properties jointly. Drawing on the professional experience of an architect on our author team, we formulate four requirements: \req{1}~\emph{design in context}, showing the proposal within its surroundings; \req{2}~\emph{architectural
plausibility}, proposing unseen sides in the language of the visible design; \req{3}~\emph{geometric consistency}, maintaining one coherent scene across views; and \req{4}~\emph{orbit completeness}, revealing
the side and rear views and returning to the start. %

Our experiments show that off-the-shelf models for video generation struggle to meet these
requirements jointly (\cref{sec:experiments}). Base video models often
produce limited camera motion, failing \req{4}. Camera-controlled models~\cite{ren2025gen3c,sun2024dimensionx,zhou2025stablevirtualcamera} move the camera, but can produce implausible unseen sides~(\req{2}), change the building's shape across views~(\req{3}), or stop partway around~(\req{4}).
Object-centric 3D generators such as TRELLIS.2~\cite{xiang2025trellis2} produce assets that can be rendered along closed orbits, but do not jointly generate the surrounding site~(\req{1}).

We hypothesize that limited training supervision for complete architectural orbits contributes to these failures. Capturing such
orbits can be difficult, especially at eye level, where neighboring buildings, vegetation, and access constraints may obstruct the camera path. Existing 3D asset and scene-video datasets are not curated
specifically for closed orbits around individual buildings with their surroundings~\cite{objaverseXL,zhou2018stereo,Ling_2024_CVPR}.

The failures of off-the-shelf video models motivate the need for a new approach designed to satisfy the requirements of ArchVis.
We introduce \emph{DesignOrbit}, a training approach guided by the four requirements (\cref{fig:method}). Procedurally generated buildings~\cite{Tsai20253D} provide a source of known 3D geometry (\req{2}) and complete camera orbits (\req{4}). We add photorealistic surroundings through a novel split-and-condition video inpainting approach that preserves
loop closure and background parallax (\req{1}), then train an initial model on these contextual orbits. Although the procedural buildings are biased toward cuboid shapes, they provide an initialization that we refine
using the model's own generations. Structure from motion (SfM)~\cite{schoenberger2016sfm} on the model's own creations allow us to assess geometry consistency~(\req{3}) and measure orbit coverage~(\req{4}) from the recovered camera poses. These measurements allow us to
select generations for self-training and further finetune for our design goals using direct preference optimization
(DPO)~\cite{rafailov2023direct,wallace2023diffusiondpo}.

We evaluate our trained model on the same requirements that motivate the design, and compare to existing video and 3D generative models.
On architecture, mean orbit coverage (\req{4}) rises from $28.77^\circ$ for the base model to $345.29^\circ$, with $86.20\%$ of videos reaching at least $351^\circ$, against $303.55^\circ$ and $25.00\%$ for the strongest baseline. The model also generates orbits for sketches, playgrounds, vessels and vehicles, interiors, and urban scenes without
further fine-tuning. To assess qualities that reconstruction alone
cannot establish, including design context~(\req{1}) and architectural
plausibility~(\req{2}), we conduct a user study. DesignOrbit was preferred over DimensionX, GEN3C, Stable Virtual Camera,
and TRELLIS.2 in $91.4\%$ of ratings that expressed a preference
(\cref{sec:user_study}).

Our contributions are as follows:
\begin{itemize}
  \item We formulate four requirements for concept-stage orbital ArchVis and analyze the limitations of current video and 3D
  generative models in meeting them jointly.
  \item Guided by these requirements, we propose a training approach
  combining synthetic data, geometrically verified self-training, and coverage-ranked preference optimization, using synthetic data.
  \item We introduce a split-and-condition video inpainting approach
  that places isolated 3D buildings in context while preserving endpoint closure and geometry consistency.
  \item We demonstrate that DesignOrbit is preferred by users on every requirement and reaches at least $351^\circ$ of orbit coverage $>3$ times as often as the strongest prior method.
\end{itemize}

\begin{figure*}[t]
  \centering
  \includegraphics[width=\textwidth, trim=0 0 0 0, clip]{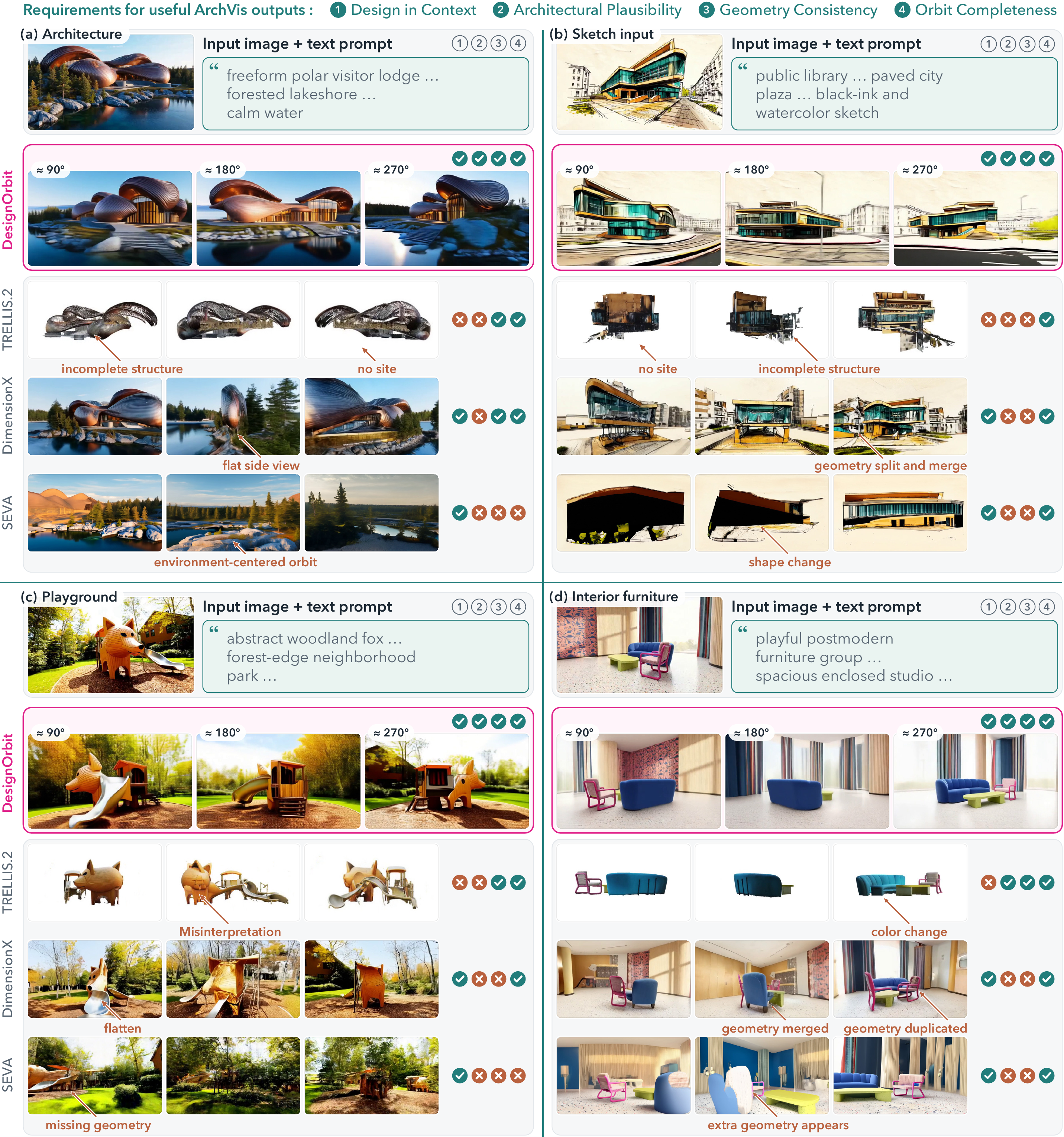}
  \setlength{\abovecaptionskip}{1pt}
\caption{\textbf{From one image and text prompt, \method{} generates 360$^\circ$ design orbits satisfying all four ArchVis requirements (Sec.~\ref{sec:background}).}
Circles indicate which requirements each output meets; every baseline fails at least one.
Although trained only on synthetic buildings, \method{} generalizes to other design-in-context examples (c,d).
TRELLIS.2 lacks surroundings and sometimes misinterprets or incompletely reconstructs the design.
DimensionX flattens, splits, or duplicates geometry, while Stable Virtual Camera (SEVA) changes geometry and sometimes fails to complete the orbit (a,c).}
  \label{fig:not_teaser}
\end{figure*}

\section{Related Work}
\label{sec:related}

\subsection{Architectural visualization}
\label{sec:related_archvis}

\noindent\textbf{ArchVis in practice.}
Architectural visualization (ArchVis) lets architects and clients evaluate
a proposed design before it is built~\cite{AIA}, and the choice of
representation affects its perceived credibility among architects, other
professionals, and members of the
public~\cite{batesbrkljac2009credibility}. Visualization is particularly
valuable during early design, while changes remain comparatively
inexpensive~\cite{paulson1976designing,braganca2014early}, but conventional
3D production~\cite{borkowski2023rendering} makes animation costly to
repeat as a concept changes~\cite{cadcrowd_animation_costs}. Designers have begun to use generative images for
ideation~\cite{paananen2024ideation,chaos2024stateofai}, which show only
one view. We address the remaining gap, seeing an early concept from every
side in its setting, and formulate requirements for evaluating such
visualizations (\cref{sec:requirements}).

\subsection{Generative models for novel views}
\label{sec:related_generative}

\noindent\textbf{3D assets and object-orbit synthesis.}
TRELLIS.2 and Hunyuan3D-2 generate textured 3D
assets~\cite{xiang2025trellis2,hunyuan3d22025tencent}, while SV3D and
Hi3D generate object-orbit views for subsequent
reconstruction~\cite{voleti2024sv3d,yang2024hi3d}.
Wang et al.~\cite{wang2026orbital} condition orbital video generation
on latent features from a 3D foundation model.
All of these treat the object in isolation, whereas DesignOrbit generates
the design and its site jointly~(\req{1}).

\noindent\textbf{Scene and camera-controlled generation.}
Multi-view diffusion models such as CAT3D synthesize novel views of a
scene for subsequent reconstruction~\cite{gao2024cat3d}.
To produce views along a particular trajectory, prior work either trains video generators to take cameras as a condition
~\cite{sun2024dimensionx,zhao2024genxd, zhou2025stablevirtualcamera}, or injects camera parameters into
pretrained video models~\cite{he2024cameractrl,wang2024motionctrl}, or specifies the camera trajectory using an input video~\cite{bai2025recammaster}.
Related approaches guide generation with explicit
3D representations, be they point
clouds~\cite{yu2024viewcrafter,ren2025gen3c,Yu_2025_ICCV} or Gaussian splats~\cite{fan2026orbitforgetextto3dscenegeneration}.
All of these control the camera at inference, through pose conditioning or
rendered views of a point cloud, and leave the content of the unseen side
to the pretrained prior. DesignOrbit instead changes the training
supervision, so that the model learns complete, consistent
orbits~(\req{3},~\req{4}) with plausible unseen sides~(\req{2}), and takes no camera input at inference.

\noindent\textbf{Training data.}
Objaverse and Objaverse-XL provide renderable
assets~\cite{objaverse,objaverseXL}, but as isolated objects without a
site. CO3D and uCO3D provide real object-centric captures, with uCO3D
targeting full $360^\circ$
coverage~\cite{reizenstein2021co3d,liu2025uco3d}, but of everyday objects
rather than buildings.
Scene datasets either have unordered image collections~\cite{li2018megadepth,tung2024megascenes} or videos where the viewpoint varies in height and distance within a
video, rather than following a closed orbit at a fixed height and
distance around a single
building~\cite{zhou2018stereo,Ling_2024_CVPR}. None of these offers
loop-closed orbit videos of buildings with their surroundings, let alone
with the design variety that exploration requires, so DesignOrbit
constructs this supervision synthetically (\cref{sec:method}).

\noindent\textbf{Geometry-guided model improvement.}
VFusion3D trains a 3D generator on video-generated multiview data
filtered by a learned quality classifier~\cite{han2024vfusion3d}.
We instead use SfM-based geometric verification to select generated
videos for self-training the orbit generator.
Diffusion-DPO adapts preference optimization to diffusion
models~\cite{wallace2023diffusiondpo}, while Epipolar-DPO, VideoGPA,
and VIGOR use geometric consistency signals to improve video
generation~\cite{kupyn2025epipolar,du2026videogpa,yin2026vigor}.
These consistency objectives do not explicitly reward complete orbits~(\req{4}). DesignOrbit uses SfM-measured registration
and coverage to select self-training videos, and coverage to rank
preference pairs.

\label{sec:related_vision_arch}

\noindent\textbf{Generative models for architecture.}
Prior vision and graphics work on architecture generates several
representations of a design: 3D building models using procedural
geometry~\cite{muller2006procedural,Tsai20253D}, floor
plans~\cite{nauata2020house,shabani2023housediffusion}, and city-scale
scenes~\cite{lin2023infinicity,xie2024citydreamer}.
Text-to-image generation and sketch-based spatial
conditioning~\cite{rombach2022highresolution,zhang2023adding} also
support architectural concept visualization
~\cite{paananen2024ideation,lee2024generative}, producing individual
views of a design.
Our task is different: given a single architectural concept image,
DesignOrbit generates alternative completions of the unseen sides,
each presented as a full orbit within its surroundings.

\begin{figure*}[t]
  \centering
  \includegraphics[width=\textwidth]{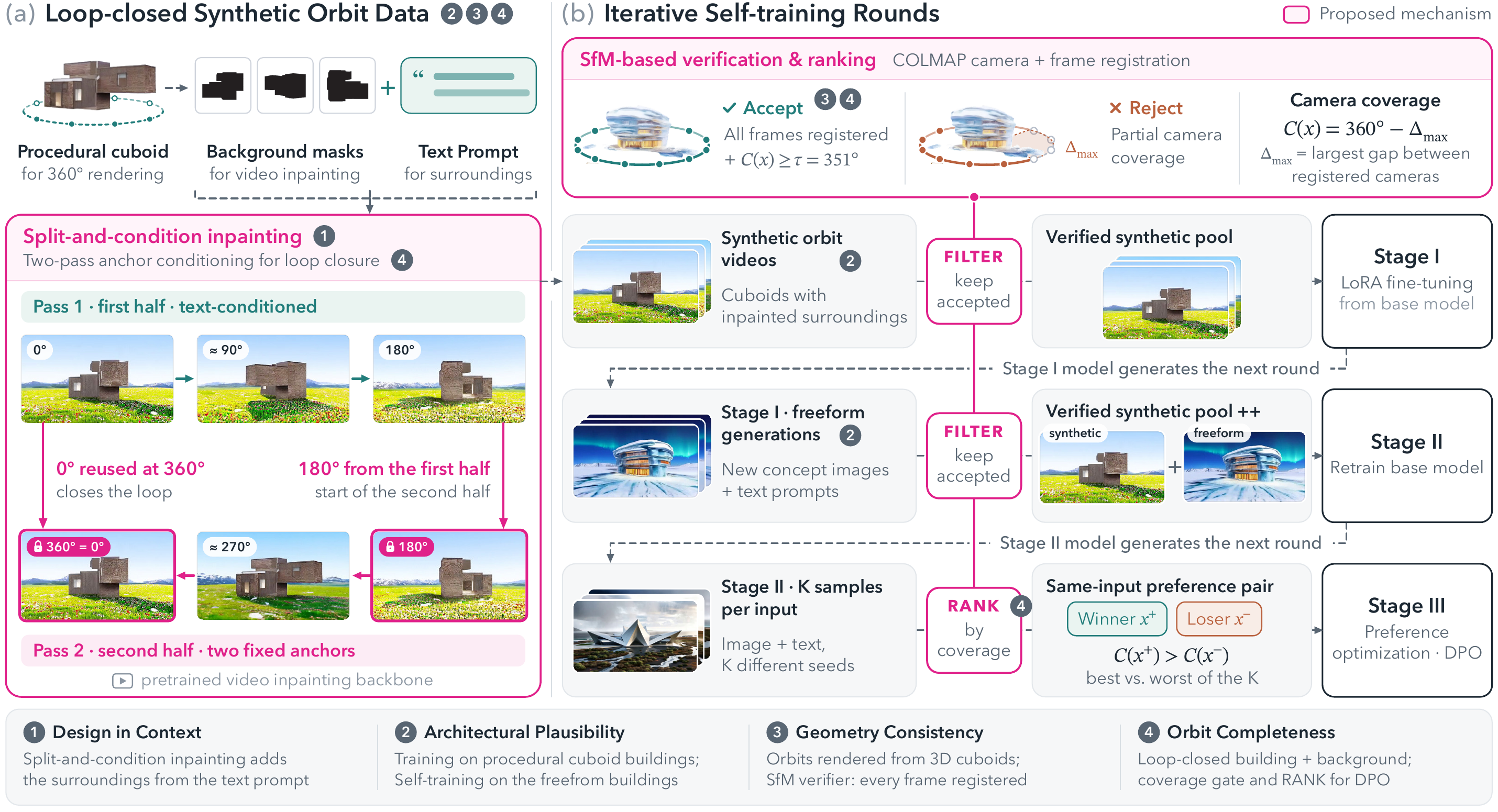}
\caption{\textbf{\method{} trains on synthetic loop-closed orbits, then uses SfM verification to filter and rank its own generations.}
(a) Procedural cuboids~\citep{Tsai20253D} are rendered along $360^\circ$ orbits, with surroundings inpainted in two passes to ensure loop closure (Sec.~\ref{sec:split-and-condition}).
(b) The verifier checks full-frame registration and coverage $C(x)\geq351^\circ$ (Sec.~\ref{sec:verifier}).
Stage~I trains on verified synthetic videos; Stage~II adds verified generations; Stage~III ranks generations by coverage for DPO (Sec.~\ref{sec:dpo}).
\req{1}--\req{4} indicate the ArchVis requirements addressed by each component.}

  \label{fig:method}
\end{figure*}

\section{Background}
\label{sec:background}

Given a concept image $I_0$, a prompt $c$ describing the design and its
surroundings, and random noise $\xi$, we seek a generator whose output
$x=G(I_0,c,\xi)$ depicts a complete $360^\circ$ orbit of one coherent design
within its setting, returning to its starting viewpoint. Different noise
should propose plausible alternatives for unseen parts while preserving the
design specified by the input.

\subsection{Requirements for useful ArchVis outputs}
\label{sec:requirements}

Drawing on the professional experience of an architect on our author team,
we identify four requirements for a generation to be useful in architectural
design, reflecting how concept visualizations are evaluated in client
presentations and design juries. They serve as design goals for the method
(\cref{sec:method}) and define the evaluation (\cref{sec:experiments}).

\noindent\textbf{\req{1} Design in context.}
The video shows the design together with its site, because a building is
judged by how it meets its setting. The surroundings must move with correct
parallax and return to the start, so we build context into the training
data (\cref{sec:synthetic}).

\noindent\textbf{\req{2} Architectural plausibility.}
The sides that $I_0$ does not show are completed with depth, massing, and
materials that continue the language of the visible side. No ground truth
exists for them, so we learn this prior from procedural buildings and widen
it by self-training (\cref{sec:synthetic,sec:method}).

\noindent\textbf{\req{3} Geometry consistency.}
All frames depict one static 3D scene: the design and its surroundings must
not flatten, change shape, duplicate, or split and merge as the camera
moves. We measure this by SfM registration (\cref{sec:verifier}).

\noindent\textbf{\req{4} Orbit completeness.}
The camera travels the full $360^\circ$ and closes the loop; a partial orbit
leaves sides unseen. We build loop closure into the data and the
conditioning, and optimize measured coverage (\cref{sec:dpo}).

These requirements are hard to meet jointly. Consistency~\req{3} is easiest
when the camera barely moves, which works against completeness~\req{4}. And
a consistent, complete orbit is far easier for an isolated object than for
a design in its site~(\req{1}), where the surroundings must also move with
correct parallax and return to the start.

\subsection{Contrast with general-purpose video models}
\label{sec:contrast}

General-purpose video models learn from broad image and video
collections~\cite{wan2025}; their standard pretraining objectives do not
explicitly enforce the four requirements above.
In our evaluations, camera-controlled video models often distort
geometry or leave orbits incomplete, while 3D generators lack
surrounding context (\cref{sec:experiments}).
These failures suggest that camera control or a renderable 3D asset
alone is insufficient to satisfy all four requirements jointly.
We posit that video generative models can better satisfy them through training data and objectives designed to address them together.

\section{Method}
\label{sec:method}

We propose an approach to fine-tune existing video generative models to
satisfy the requirements of \cref{sec:requirements}, combining synthetic and model-generated videos.

Our overall pipeline is as follows (\cref{fig:method}):
\begin{enumerate}
  \item In \textbf{Stage~I}, we use a procedural approach to produce a
  dataset of synthetic orbit videos (\cref{sec:synthetic}), and then filter them using our SfM-based verification (\cref{sec:verifier}). On this dataset of approximately 300 verified clips, we fine-tune
  Wan2.1-FLF2V-14B~\cite{wan2025}. We use the same image for both endpoint
  conditions, so that loop closure~(\req{4}) is a condition given to the
  model and matches the loop-closed training clips.
  \item Our synthetic data is biased towards cuboidal buildings because of
  the limitations of the procedural approach. In \textbf{Stage~II:
  self-training beyond cuboids}, we use the Stage~I model to generate orbits of free-form building concepts and retain approximately 300 clips after SfM-based verification. We combine these with the synthetic training set and train a fresh LoRA from the original pretrained Wan checkpoint on
  approximately 600 clips. This extends architectural
  plausibility~\req{2} to free-form designs, while the verifier guarantees that the added clips retain \req{3} and \req{4}.
  \item Finally, \textbf{Stage~III} applies Direct Preference Optimization (DPO)~\cite{wallace2023diffusiondpo} to push generations toward complete
  orbits~(\req{4}): for each input we draw $K$ samples, rank them by SfM
  camera coverage, and form winner--loser pairs from the same input
  (\cref{sec:dpo}).
\end{enumerate}

Each component targets specific requirements (\cref{fig:method}). Procedural building models rendered along closed circular camera paths provide architectural plausibility~\req{2}, geometry consistency~\req{3}, and orbit
completeness~\req{4} for the building; split-and-condition inpainting adds
the context~\req{1} and extends \req{3} and \req{4} to the surroundings. The resulting training data is thus designed to satisfy all four requirements.
SfM verification then measures \req{3} and \req{4} on model-generated videos, which lets us select them for self-training, extending plausibility~\req{2} beyond procedural shapes, and rank them for preference
optimization, further enforcing \req{4}. The verifier does not measure
\req{1} and \req{2}; these enter through the training data, and we check in the user study that the later stages do not erode them.

In the subsections below, we delve deeper into the key technical
contributions. Sampling budgets, acceptance rates, and hyperparameter
settings are provided in the supplementary material.

\subsection{Synthetic orbits with contextual backgrounds}
\label{sec:synthetic}

We bootstrap self-training with synthetic data because none of the
approximately 300 videos sampled from the base pretrained video generation
model passed our SfM verifier. To create this data, we first use an LLM to generate prompts describing buildings and their settings and then use these prompts to generate procedural cuboid buildings
following~\cite{Tsai20253D}. We then render these procedural generations
along known circular camera paths with identical start and end views.
Because every frame is rendered from a single procedural 3D building model,
the clips are geometrically consistent~(\req{3}) and show architecturally
plausible side and rear views~(\req{2}) by construction, and the closed camera path guarantees a complete orbit~(\req{4}).

The rendered clips have no background, and hence no context~(\req{1}).
Training on them alone produces buildings that rotate against a static
background (\cref{sec:ablation}). To make the background rotate with the building, we mask out everything except the building in every frame and
inpaint the masked regions using a video inpainting model~\cite{vace}. The inpainting prompt is generated by an LLM and describes only the surrounding environment.

\textbf{Split-and-condition inpainting.}
\label{sec:split-and-condition}
In our experiments, unconstrained single-pass inpainting fails to close the loop: the beginning of the video is inconsistent with the end. The
alternative of conditioning the whole sequence on identical endpoint images
can produce a static background. Instead, we split the orbit at
$m=\lfloor (T-1)/2 \rfloor$ and first inpaint frames $0{:}m$, conditioned on
the rendered foreground and setting prompt. We then inpaint frames
$m{:}T{-}1$ using the completed midpoint and first frame as the starting and ending anchors, respectively, protecting both anchors and the foreground with masks. This retains the first half's camera-induced background motion while joining the second half to both anchors, so that the inpainted
context~(\req{1}) is itself a complete, closed orbit~(\req{4}).

In spite of these measures, our approach might still introduce repeated
textures or unwanted structures attached to the building. We therefore
filter the completed clips with the verifier below, which discards clips
that fail COLMAP SfM reconstruction and the camera-coverage check, and
retain approximately 300 as $\mathcal{D}_\text{syn}$. Qualitative examples appear in the supplementary material.

\subsection{SfM verification of consistency and coverage}
\label{sec:verifier}

Our training pipeline relies on an SfM-based verifier both to filter bad
videos from the training set and to provide preference pairs for
optimization. The verifier checks the two requirements: the consistency of geometry across the video~(\req{3})
and the completeness of the orbit~(\req{4}). It works as follows.

For each candidate video $x$, we use COLMAP~\cite{schoenberger2016sfm} to
estimate cameras and a sparse point cloud. We check that the number of
registered frames, $N_\text{reg}(x)$, is equal to the length of the video,
$T=81$. This supplies evidence of geometric consistency~(\req{3}), but an
incomplete orbit can also register successfully.

To measure coverage~(\req{4}), we fit an orbital plane to the recovered
camera centers by SVD, estimate the orbit center, and compute camera
azimuths. We remove cameras that are inconsistent in travel direction or
radius using $3\times$MAD outlier filtering (details in the supplementary
material). For the retained, azimuth-sorted cameras, let $\Delta_i$ denote
consecutive angular gaps, including the gap between the last and the first.
We define orbit coverage as $C(x)=360^\circ-\max_i\Delta_i$.
Finally, we consider a video as verified,
$V(x)=1$, if (a) the number of registered frames is equal to the length of
the video, and (b) the coverage is at least $\tau=351^\circ$, corresponding
to at most a single missed frame.\footnote{For an ideal 81-frame orbit with
a repeated endpoint, the spacing is $4.5^\circ$ and $C(x)=355.5^\circ$.}
Both checks are needed: self-training with registration alone leaves the
model worse than before (\cref{sec:ablation}).

\begin{algorithm}[t]
\caption{Coverage-ranked DPO (Stage III)}
\label{alg:dpo}
\footnotesize
\begin{algorithmic}[1]
\Require Stage-II model $G_{\mathrm{II}}$, inputs $\mathcal P$,
$K=8$, $\delta=30^\circ$
\State $S(x)\gets C(x)$ if computable, else $0^\circ$
\State $\mathcal D_{\mathrm{pref}}\gets\emptyset$
\For{$p\in\mathcal P$}
    \State $\mathcal X\gets\{G_{\mathrm{II}}(p;\xi_k)\}_{k=1}^{K}$
    \State $x^+,x^-\gets
    \arg\max_{x\in\mathcal X}S(x),
    \arg\min_{x\in\mathcal X}S(x)$
    \If{$S(x^+)-S(x^-)\geq\delta$}
        \State $\mathcal D_{\mathrm{pref}}\gets
        \mathcal D_{\mathrm{pref}}\cup\{(p,x^+,x^-)\}$
    \EndIf
\EndFor
\State Split $\mathcal D_{\mathrm{pref}}$ by input into
$\mathcal D_{\mathrm{train}},\mathcal D_{\mathrm{val}}$
\State \Return $\textsc{DPO}(G_{\mathrm{II}},\mathcal D_{\mathrm{train}})$
\end{algorithmic}
\end{algorithm}

\subsection{Coverage-ranked preference optimization}
\label{sec:dpo}

The binary filtering above discards many bad videos, but bad videos can
serve as useful \emph{negative} examples. Concretely, $90\%$ of the videos generated by our Stage~II model have all their frames registered during SfM, suggesting that they are geometrically consistent~(\req{3}). However, many of them may not be complete orbits. We therefore rank videos by $C(x)$, which measures the viewing directions available for inspection, and not by a consistency score, which would favor less camera motion (\cref{sec:requirements}). Ranking-criteria ablations are in the supplementary material.

For each of 360 image--text inputs $p=(I_0,c)$, we sample $K=8$ Stage~II
videos. We set $S(x)=C(x)$ when coverage is computable and $S(x)=0^\circ$ otherwise. Within each input, we pair the highest-scoring video $x^+$ with the lowest-scoring video $x^-$ and retain the pair if
$S(x^+)-S(x^-)\ge 30^\circ$. This yields 350 pairs, one per retained input. We reserve approximately $15\%$ of these inputs for validation and train on the remaining pairs (\cref{alg:dpo}).

We adapt direct preference optimization~\cite{rafailov2023direct} and Diffusion-DPO~\cite{wallace2023diffusiondpo} to the model's flow-matching
parameterization~\cite{lipman2023flowmatching}. For each pair $(p,x^+,x^-)$, we encode videos as latents $z^\pm$ and draw a shared Gaussian noise sample
$\epsilon\sim\mathcal{N}(0,I)$ and noise fraction $t$, forming
$z^\pm_t=(1-t)z^\pm+t\epsilon$ with velocity targets $u^\pm=\epsilon-z^\pm$.
For each video, the policy and the frozen Stage~II reference model process
the same noisy latents. We define
\begin{equation}
  e^\pm_\theta = \mathrm{MSE}\!\left(v_\theta(z^\pm_t,t,p),\,u^\pm\right),
  \qquad
  d^\pm_\theta = e^\pm_\theta - e^\pm_\text{ref},
\end{equation}
where MSE averages over latent elements and $e^\pm_\text{ref}$ is the
reference error. We minimize
\begin{equation}
  \mathcal{L} = \mathbb{E}_{(p,x^+,x^-),\,t,\,\epsilon}
  \left[-\log\sigma\!\left(-\beta\,(d^+_\theta-d^-_\theta)\right)
  + \lambda\, e^+_\theta\right],
  \label{eq:dpo}
\end{equation}
where $\sigma$ is the logistic function. 
The preference term favors greater error reduction on winners
relative to losers, while the auxiliary SFT term, when enabled,
encourages low absolute error on winners. Our ablation shows
that $\lambda=0$ yields higher orbit coverage than $\lambda=0.1$; we therefore adopt the former in our final model. We merge the Stage~II LoRA into the base weights, freeze them, and train a fresh adapter initialized to produce zero weight updates. 
We sample $t$ at high noise levels; the exact sampling rule and training hyperparameters are in the supplementary material.

After training, coverage rises from $318.21^\circ$ to $349.04^\circ$
(\cref{tab:ablation_main}). Because this stage optimizes \req{4} only, we check that it does not cost the other requirements: study participants still rate the unseen sides as compatible with the input and as usable design alternatives (\cref{tab:user_study}). 

\noindent\textbf{Inference.}
Given a sketch or concept image and its prompt, DesignOrbit directly samples orbit videos without requiring a pre-existing 3D model or additional views. We sample multiple candidate completions by varying the seed.

\section{Experiments}
\label{sec:experiments}

\begin{table}[t]
\centering
\caption{\textbf{User study (50 participants): \method{} is preferred across all six criteria, receiving 91\% of expressed preferences.}
Entries show \method{} / baseline preferences; remaining ratings are ties or neither.
Columns 1--3 compare single videos (100 ratings per baseline); 4--6 compare video pairs (50 per baseline).
}

\label{tab:user_study}
\label{tab:user_study_baseline}

\footnotesize
\setlength{\tabcolsep}{1.5pt}
\renewcommand{\arraystretch}{1.1}

\begin{tabular*}{\columnwidth}
{@{\extracolsep{\fill}}lcccccc@{}}
\toprule
& \multicolumn{3}{c}{Single video}
& \multicolumn{3}{c}{Video pairs} \\
\cmidrule(lr){2-4}
\cmidrule(l){5-7}

Method
& \shortstack{Present.\\pref.}
& \shortstack{Shape\\consis.}
& \shortstack{Views\\compat.}
& \shortstack{Views\\visible}
& \shortstack{Unseen\\variety}
& \shortstack{Usable\\alter.} \\
\midrule

DimensionX
& \textbf{68}/6
& \textbf{71}/1
& \textbf{68}/2
& \textbf{34}/6
& \textbf{52}/10
& \textbf{48}/10 \\

SEVA
& \textbf{66}/7
& \textbf{70}/0
& \textbf{55}/5
& \textbf{54}/4
& \textbf{58}/16
& \textbf{68}/4 \\

GEN3C
& \textbf{74}/5
& \textbf{80}/3
& \textbf{67}/9
& \textbf{78}/2
& \textbf{62}/8
& \textbf{76}/2 \\

TRELLIS (white)
& \textbf{78}/7
& \textbf{28}/12
& \textbf{39}/11
& \textbf{38}/2
& \textbf{38}/18
& \textbf{46}/12 \\

TRELLIS (inpainted)
& \textbf{86}/4
& \textbf{65}/1
& \textbf{66}/3
& --
& --
& -- \\

\midrule

All baselines
& \textbf{74}/6
& \textbf{63}/3
& \textbf{59}/6
& \textbf{51}/4
& \textbf{53}/13
& \textbf{60}/7 \\

\bottomrule
\end{tabular*}
\vspace{-10pt}
\end{table}

\textbf{Metrics.}
Our experiments test each of the four ArchVis requirements
(\cref{sec:requirements}) against the video models a designer could use today.
Geometry consistency~\req{3} and orbit completeness~\req{4} are measured
using SfM verifier of Section~\ref{sec:verifier}, while context~\req{1} and
plausibility~\req{2} are judged by people
(Section~\ref{sec:user_study}). We also test whether the requirements are
met beyond the training domain and for sketch inputs.

\subsection{Setup}
\label{sec:setup}

\noindent\textbf{Data.}
We train only on architectural scenes and evaluate on two categories: \emph{Architecture} (in-domain) and \emph{Other design-in-context} (out-of-domain), including public art, playgrounds, interiors, vehicles, and imagined environments. The user study additionally includes architectural sketches to test generalization across
input representations.

\noindent\textbf{Inputs and protocol.}
We use Qwen2.5-VL-7B-Instruct~\cite{qwen2.5} to generate prompts and Wan2.1-T2V-14B~\cite{wan2025} to generate concept images from the first frames of videos. All methods receive identical
image--text inputs: $\sim$190 for Architecture (19 prompts, 10 images each) and $\sim$210 for Other design-in-context.
Test prompts are distinct from training prompts, and all test images are newly generated. For baseline comparison, all videos are standardized to 81 frames, 15\,fps, $480\times320$.

\noindent\textbf{Baselines.}
We compare four families of models:
(i)~\emph{Camera-controlled video} targeting orbit motion~\req{4}: GEN3C~\citep{ren2025gen3c},
ViewCrafter~\citep{yu2024viewcrafter},
TrajectoryCrafter~\citep{Yu_2025_ICCV},
ReCamMaster~\citep{bai2025recammaster}, and Stable Virtual Camera~\citep{zhou2025stablevirtualcamera};
(ii)~\emph{Reconstruction-oriented video} targeting geometry~\req{3}: DimensionX~\citep{sun2024dimensionx}
and GenXD~\citep{zhao2024genxd};
(iii)~\emph{Single-image 3D generation} with fixed geometry and renderable viewpoints~(\req{3}, \req{4}) but no surroundings~(\req{1}): TRELLIS.2~\citep{xiang2025trellis2}
and Hunyuan3D-2~\citep{hunyuan3d22025tencent}.
We also evaluate TRELLIS.2 with inpainted surroundings.
(iv)~\emph{Base video models} without orbit-specific training, which fail \req{4}: Wan2.1-FLF2V-14B, Wan2.1-I2V-14B, and HunyuanVideo-I2V-13B (results in supplementary).
All baselines use released code and recommended settings.

\subsection{User Study}
\label{sec:user_study}

We recruited 50 participants, including 40 with architecture
or design experience, to compare \method{} against GEN3C, Stable Virtual Camera, DimensionX, and TRELLIS.2 (white and inpainted backgrounds). Inputs span Architecture,
Other design-in-context, and architectural sketches.
In \emph{single-video comparisons} (100 ratings per baseline), participants judged presentation in context~\req{1}, shape consistency~\req{3}, and compatibility of unseen views~\req{2}. In \emph{video-set comparisons}
(50 ratings per baseline, excluding inpainted TRELLIS.2), they evaluated visibility of sides and back~\req{4},
variety in unseen parts, and usability of design alternatives.

\noindent\textbf{Participants prefer \method{} across all six criteria} (Table~\ref{tab:user_study}), choosing it in 91\% of ratings expressing a preference. TRELLIS.2 on a white background comes closest on shape
consistency~\req{3}, but falls further behind once surroundings are inpainted, highlighting the difficulty
of satisfying context~\req{1} and consistency~\req{3} together. Preferences extend beyond Architecture, with the largest margins on sketches (see supplementary). Although margins are smallest for unseen variety, \method{} remains preferred, and diversity is maintained
across training stages (see supplementary).

\subsection{Quantitative Evaluation}
\label{sec:quantitative}

We evaluate orbit completeness~\req{4} and geometric consistency~\req{3} against seven video baselines
(\cref{sec:setup}). Orbit completeness requires all 81 frames to register in COLMAP with camera coverage $C(x)\geq351^\circ$ (\cref{sec:verifier}).
For consistency, we count reconstructed foreground 3D points, identified using SAM~2 masks~\cite{ravi2024sam2}.

\noindent\textbf{\method{} achieves substantially more complete orbits~\req{4}} (Table~\ref{tab:cross_model_compare1}).
On Architecture, \method{} completes 86.2\% of orbits, compared with 25.0\% for the strongest baseline, Stable
Virtual Camera (SEVA). The advantage extends to other design-in-context inputs (69.8\% vs.\ 20.5\%).
Even DimensionX, with mean coverage above $300^\circ$, completes only 6.3\% of architectural orbits.
\method{} also reconstructs the most foreground points (1897), providing additional evidence of geometric consistency~\req{3}.

\noindent\textbf{Baselines fail in different ways.}
DimensionX flattens unseen sides, while SEVA and GEN3C change geometry during the orbit~(\req{3}).
ReCamMaster and ViewCrafter leave orbits incomplete~(\req{4}), whereas GenXD and TrajectoryCrafter alter the design and its surroundings~(\req{1}, \req{3}). DimensionX retains a relatively high point count despite
flattening geometry, illustrating that reconstructability alone does not guarantee shape fidelity. Together, these results show that \method{} better combines orbit completeness~\req{4}, geometric
consistency~\req{3}, and design in context~\req{1}.

\begin{table}[t]
\centering
\caption{\textbf{\method{} achieves over $3\times$ higher full-orbit rates than the strongest baseline, with each training component contributing to completeness.}
(a) Comparison with video baselines.
(b) Key training ablations on architectural inputs.
Full: percentage of videos with all 81 frames registered and
$C(x)\geq351^\circ$; Cov.: mean coverage ($0^\circ$ if SfM fails);
Pts: mean foreground COLMAP points.
$\sim$190/$\sim$210 videos per method for architecture/other.}
\label{tab:combined_results}

\footnotesize
\setlength{\tabcolsep}{2pt}
\renewcommand{\arraystretch}{1.08}

\begin{subtable}[t]{\columnwidth}
\centering
\caption{Comparison with video baselines $480\times320$ ($W\times H$)}
\label{tab:cross_model_compare1}

\begin{tabular*}{\linewidth}
{@{\extracolsep{\fill}}lccccc@{}}
\toprule
& \multicolumn{2}{c}{Architecture}
& \multicolumn{2}{c}{Other}
& \\
\cmidrule(lr){2-3}
\cmidrule(lr){4-5}
Method
& Full $\uparrow$
& Cov. $\uparrow$
& Full $\uparrow$
& Cov. $\uparrow$
& Pts $\uparrow$ \\
\midrule

\textbf{\method{}}
& \textbf{86.20\%}
& \textbf{345.29$^\circ$}
& \textbf{69.80\%}
& \textbf{332.35$^\circ$}
& \textbf{1897} \\

DimensionX
& 6.30\% & 302.40$^\circ$
& 11.10\% & 316.62$^\circ$ & 1792 \\

GenXD
& 0.00\% & 75.36$^\circ$
& 0.00\% & 0.00$^\circ$ & 8 \\

SEVA
& 25.00\% & 303.55$^\circ$
& 20.50\% & 309.81$^\circ$ & 1450 \\

GEN3C
& 0.50\% & 200.22$^\circ$
& 0.90\% & 179.45$^\circ$ & 818 \\

ReCamMaster
& 0.00\% & 125.75$^\circ$
& 0.00\% & 107.64$^\circ$ & 743 \\

ViewCrafter
& 0.00\% & 123.52$^\circ$
& 0.50\% & 152.87$^\circ$ & 712 \\

TrajectoryCrafter
& 0.00\% & 93.60$^\circ$
& 0.00\% & 122.82$^\circ$ & 437 \\

\bottomrule
\end{tabular*}
\end{subtable}

\vspace{6pt}

\begin{subtable}[t]{\columnwidth}
\centering
\caption{Key training ablations $832\times480$ ($W\times H$)}
\label{tab:ablation_main}

\setlength{\tabcolsep}{3pt}

\begin{tabularx}{\linewidth}{@{}Xcc@{}}
\toprule
Variant
& Full $\uparrow$
& Cov. $\uparrow$ \\
\midrule

\multicolumn{3}{@{}l}{\textit{(i) Data construction (Stage I)}} \\

\textbf{Procedural + loop-closed inpainting}
& \textbf{11.70\%}
& \textbf{301.15$^\circ$} \\

Single-pass inpainting
& 0.00\% & 95.19$^\circ$ \\

Without last-frame conditioning
& 9.60\% & 298.01$^\circ$ \\

Raw renders, no inpainting
& 0.00\% & 70.74$^\circ$ \\

\midrule
\multicolumn{3}{@{}l}{\textit{(ii) SfM verification (Stage II)}} \\

\textbf{Registration + coverage}
& \textbf{16.50\%}
& \textbf{318.21$^\circ$} \\

Registration only
& 3.70\% & 288.93$^\circ$ \\

No verification
& 0.00\% & 250.86$^\circ$ \\

\midrule
\multicolumn{3}{@{}l}{\textit{(iii) Optimization (Stage III)}} \\

\textbf{Coverage-ranked DPO}
& \textbf{92.60\%}
& \textbf{349.04$^\circ$} \\

SFT with two-filter verification
& 40.40\% & 332.38$^\circ$ \\

\bottomrule
\end{tabularx}
\end{subtable}
\vspace{-10pt}
\end{table}

\subsection{Ablation Studies}
\label{sec:ablation}

We ablate key design decisions for useful ArchVis outputs
(Table~\ref{tab:ablation_main}), measuring orbit
completeness~\req{4} on $\sim$190 architectural inputs.
All variants use Wan2.1~\cite{wan2025} at $832\times480$, different from the
resolution in Table~\ref{tab:cross_model_compare1}.

\noindent\textbf{Stage I: loop-closed data~(\req{1}, \req{4}).}
Training on raw renders without inpainting yields no complete orbits and only $70.74^\circ$ mean coverage.
Single-pass inpainting breaks the loop because the surroundings do not return to their starting view, reducing the full-orbit rate from 11.7\% to 0\% and
mean coverage from $301.15^\circ$ to $95.19^\circ$. Removing last-frame conditioning also lowers the
full-orbit rate to 9.6\% and introduces visible artifacts at the loop boundary.

\noindent\textbf{Stage II: SfM verification~(\req{3}, \req{4}).}
Self-training without verification yields no complete orbits, while registration-only filtering achieves 3.7\%. Adding coverage verification raises this to 16.5\%, showing that geometric reconstructability alone does
not guarantee orbit completeness.

\noindent\textbf{Stage III: coverage-ranked DPO~(\req{4}).}
Compared with SFT on verified clips, coverage-ranked DPO increases the full-orbit rate from 40.4\% to 92.6\%
and mean coverage from $332.38^\circ$ to $349.04^\circ$. 
Further ablations on training datasets, ranking criteria, and recipe transfer to HunyuanVideo are in the supplementary.

\section{Conclusion}
ArchVis for early-stage design places four requirements on generated videos:
\req{1} design in context, \req{2} architectural plausibility, \req{3} geometric consistency, and \req{4} orbit completeness. We showed that existing video and 3D generative models fall short of satisfying these requirements together. Guided by these requirements, we developed \method{}, which generates complete $360^\circ$ orbit videos with alternative completions of unseen sides from a single concept image or sketch
and a text prompt. Our training recipe combines loop-closed synthetic data,
SfM-verified self-training, and coverage-ranked preference
optimization. \method{} achieves over three times the full-orbit rate of the strongest baseline and is preferred by study participants across all evaluation criteria.

\textbf{Limitations.}
Our training data consists of procedural cuboid buildings and
self-generated free-form buildings, mostly in open natural surroundings, so
the model sees little fine architectural detail and few dense or cluttered sites. On inputs far from this distribution, such as historical buildings with fine ornamentation or heavily occluded sites, plausibility~\req{2} and
context~\req{1} degrade. \method{} is therefore suited to the pre-visualization of contemporary designs, where the unseen sides are not yet fixed, and a plausible completion is what the designer needs; it is not intended for heritage documentation or conservation, where the unseen sides of an existing building must be reproduced faithfully.

\section{LLM Usage Statement}
LLMs were used to assist with manuscript editing, code development, and figure preparation. Generative models used for data synthesis are described in the paper. All AI-assisted content was reviewed and validated by the authors, who take full responsibility for the final work.

\section{Acknowledgments}
We thank Noah Snavely for his insightful feedback and our user study participants for their time and contributions.

{
    \small
    \bibliographystyle{ieeenat_fullname}
    \bibliography{main}

\begin{thebibliography}{63}
\providecommand{\natexlab}[1]{#1}
\providecommand{\url}[1]{\texttt{#1}}
\expandafter\ifx\csname urlstyle\endcsname\relax
  \providecommand{\doi}[1]{doi: #1}\else
  \providecommand{\doi}{doi: \begingroup \urlstyle{rm}\Url}\fi

\bibitem[{American Institute of Architects}(2023)]{AIA}
{American Institute of Architects}.
\newblock Defining the architect's basic services, 2023.

\bibitem[{Architizer} and {Chaos}(2024)]{chaos2024stateofai}
{Architizer} and {Chaos}.
\newblock The state of {AI} in architecture.
\newblock Technical report, Architizer and Chaos, 2024.
\newblock Survey of 1,227 architects and design professionals. Accessed 2026-11-12.

\bibitem[Bai et~al.(2025)Bai, Xia, Fu, Wang, Mu, Cao, Liu, Hu, Bai, Wan, et~al.]{bai2025recammaster}
Jianhong Bai, Menghan Xia, Xiao Fu, Xintao Wang, Lianrui Mu, Jinwen Cao, Zuozhu Liu, Haoji Hu, Xiang Bai, Pengfei Wan, et~al.
\newblock Recammaster: Camera-controlled generative rendering from a single video.
\newblock \emph{arXiv preprint arXiv:2503.11647}, 2025.

\bibitem[Bates-Brkljac(2009)]{batesbrkljac2009credibility}
Nada Bates-Brkljac.
\newblock Assessing perceived credibility of traditional and computer generated architectural representations.
\newblock \emph{Design Studies}, 30\penalty0 (4):\penalty0 415--437, 2009.

\bibitem[Blattmann et~al.(2021)Blattmann, Milbich, Dorkenwald, and Ommer]{ipoke}
Andreas Blattmann, Timo Milbich, Michael Dorkenwald, and Bj{\"o}rn Ommer.
\newblock {iPOKE}: Poking a still image for controlled stochastic video synthesis.
\newblock In \emph{Proceedings of the IEEE/CVF International Conference on Computer Vision (ICCV)}, pages 14707--14717, 2021.

\bibitem[Borkowski and Nowakowski(2023)]{borkowski2023rendering}
Andrzej Borkowski and Piotr Nowakowski.
\newblock Use of applications and rendering engines in architectural design -- state-of-the-art.
\newblock \emph{Budownictwo i Architektura}, 22\penalty0 (1):\penalty0 5--14, 2023.

\bibitem[Bragan{\c{c}}a et~al.(2014)Bragan{\c{c}}a, Vieira, and Andrade]{braganca2014early}
Lu{\'i}s Bragan{\c{c}}a, Susana~M. Vieira, and Joana~B. Andrade.
\newblock Early stage design decisions: The way to achieve sustainable buildings at lower costs.
\newblock \emph{The Scientific World Journal}, 2014:\penalty0 365364, 2014.

\bibitem[Chen et~al.(2024)Chen, Siarohin, Menapace, Deyneka, Chao, Jeon, Fang, Lee, Ren, Yang, and Tulyakov]{chen2024panda}
Tsai-Shien Chen, Aliaksandr Siarohin, Willi Menapace, Ekaterina Deyneka, Hsiang-wei Chao, Byung~Eun Jeon, Yuwei Fang, Hsin-Ying Lee, Jian Ren, Ming-Hsuan Yang, and Sergey Tulyakov.
\newblock {Panda-70M}: Captioning 70m videos with multiple cross-modality teachers.
\newblock In \emph{Proceedings of the IEEE/CVF Conference on Computer Vision and Pattern Recognition (CVPR)}, pages 13320--13331, 2024.

\bibitem[Deitke et~al.(2022)Deitke, Schwenk, Salvador, Weihs, Michel, VanderBilt, Schmidt, Ehsani, Kembhavi, and Farhadi]{objaverse}
Matt Deitke, Dustin Schwenk, Jordi Salvador, Luca Weihs, Oscar Michel, Eli VanderBilt, Ludwig Schmidt, Kiana Ehsani, Aniruddha Kembhavi, and Ali Farhadi.
\newblock Objaverse: A universe of annotated 3d objects.
\newblock \emph{arXiv preprint arXiv:2212.08051}, 2022.

\bibitem[Deitke et~al.(2023)Deitke, Liu, Wallingford, Ngo, Michel, Kusupati, Fan, Laforte, Voleti, Gadre, VanderBilt, Kembhavi, Vondrick, Gkioxari, Ehsani, Schmidt, and Farhadi]{objaverseXL}
Matt Deitke, Ruoshi Liu, Matthew Wallingford, Huong Ngo, Oscar Michel, Aditya Kusupati, Alan Fan, Christian Laforte, Vikram Voleti, Samir~Yitzhak Gadre, Eli VanderBilt, Aniruddha Kembhavi, Carl Vondrick, Georgia Gkioxari, Kiana Ehsani, Ludwig Schmidt, and Ali Farhadi.
\newblock Objaverse-xl: A universe of 10m+ 3d objects.
\newblock \emph{arXiv preprint arXiv:2307.05663}, 2023.

\bibitem[Dorkenwald et~al.(2021)Dorkenwald, Milbich, Blattmann, Rombach, Derpanis, and Ommer]{cinn}
Michael Dorkenwald, Timo Milbich, Andreas Blattmann, Robin Rombach, Konstantinos~G. Derpanis, and Bj{\"o}rn Ommer.
\newblock Stochastic image-to-video synthesis using {cINNs}.
\newblock In \emph{Proceedings of the IEEE/CVF Conference on Computer Vision and Pattern Recognition (CVPR)}, pages 3742--3753, 2021.

\bibitem[Du et~al.(2026)Du, Ye, Cong, Li, Ni, Agarwal, Zhou, Li, Balestriero, and Wang]{du2026videogpa}
Hongyang Du, Junjie Ye, Xiaoyan Cong, Runhao Li, Jingcheng Ni, Aman Agarwal, Zeqi Zhou, Zekun Li, Randall Balestriero, and Yue Wang.
\newblock {VideoGPA}: Distilling geometry priors for {3D}-consistent video generation, 2026.

\bibitem[Fan and Favaro(2026)]{fan2026orbitforgetextto3dscenegeneration}
Chenrui Fan and Paolo Favaro.
\newblock Orbitforge: Text-to-3d scene generation via reconstruction-anchored video synthesis, 2026.

\bibitem[Gao* et~al.(2024)Gao*, Holynski*, Henzler, Brussee, Martin-Brualla, Srinivasan, Barron, and Poole*]{gao2024cat3d}
Ruiqi Gao*, Aleksander Holynski*, Philipp Henzler, Arthur Brussee, Ricardo Martin-Brualla, Pratul~P. Srinivasan, Jonathan~T. Barron, and Ben Poole*.
\newblock Cat3d: Create anything in 3d with multi-view diffusion models.
\newblock \emph{Advances in Neural Information Processing Systems}, 2024.

\bibitem[Han et~al.(2024)Han, Kokkinos, and Torr]{han2024vfusion3d}
Junlin Han, Filippos Kokkinos, and Philip Torr.
\newblock Vfusion3d: Learning scalable 3d generative models from video diffusion models.
\newblock \emph{European Conference on Computer Vision (ECCV)}, 2024.

\bibitem[He et~al.(2024)He, Xu, Guo, Wetzstein, Dai, Li, and Yang]{he2024cameractrl}
Hao He, Yinghao Xu, Yuwei Guo, Gordon Wetzstein, Bo Dai, Hongsheng Li, and Ceyuan Yang.
\newblock Cameractrl: Enabling camera control for text-to-video generation.
\newblock \emph{arXiv preprint arXiv:2404.02101}, 2024.

\bibitem[Hu et~al.(2022)Hu, Shen, Wallis, Allen-Zhu, Li, Wang, Wang, and Chen]{hu2022lora}
Edward~J Hu, Yelong Shen, Phillip Wallis, Zeyuan Allen-Zhu, Yuanzhi Li, Shean Wang, Lu Wang, and Weizhu Chen.
\newblock Lo{RA}: Low-rank adaptation of large language models.
\newblock In \emph{International Conference on Learning Representations}, 2022.

\bibitem[Jiang et~al.(2025)Jiang, Han, Mao, Zhang, Pan, and Liu]{vace}
Zeyinzi Jiang, Zhen Han, Chaojie Mao, Jingfeng Zhang, Yulin Pan, and Yu Liu.
\newblock Vace: All-in-one video creation and editing.
\newblock In \emph{Proceedings of the IEEE/CVF International Conference on Computer Vision}, pages 17191--17202, 2025.

\bibitem[Kerbl et~al.(2023)Kerbl, Kopanas, Leimk{\"u}hler, and Drettakis]{kerbl3Dgaussians}
Bernhard Kerbl, Georgios Kopanas, Thomas Leimk{\"u}hler, and George Drettakis.
\newblock 3d gaussian splatting for real-time radiance field rendering.
\newblock \emph{ACM Transactions on Graphics}, 42\penalty0 (4), 2023.

\bibitem[Kupyn et~al.(2025)Kupyn, Uscidda, Tintore~Gazulla, Manhardt, Tombari, and Rupprecht]{kupyn2025epipolar}
Orest Kupyn, Th{\'e}o Uscidda, Marta Tintore~Gazulla, Fabian Manhardt, Federico Tombari, and Christian Rupprecht.
\newblock Epipolar geometry improves video generation models, 2025.

\bibitem[Lee et~al.(2024)Lee, Yoo, and Cha]{lee2024generative}
Jin-Kook Lee, Youngjin Yoo, and Seung~Hyun Cha.
\newblock Generative early architectural visualizations: incorporating architect's style-trained models.
\newblock \emph{Journal of Computational Design and Engineering}, 11\penalty0 (5):\penalty0 40--59, 2024.

\bibitem[Li and Snavely(2018)]{li2018megadepth}
Zhengqi Li and Noah Snavely.
\newblock Megadepth: Learning single-view depth prediction from internet photos.
\newblock In \emph{Proceedings of the IEEE/CVF Conference on Computer Vision and Pattern Recognition (CVPR)}, pages 2041--2050, 2018.

\bibitem[Lin et~al.(2023)Lin, Lee, Menapace, Chai, Siarohin, Yang, and Tulyakov]{lin2023infinicity}
Chieh~Hubert Lin, Hsin-Ying Lee, Willi Menapace, Menglei Chai, Aliaksandr Siarohin, Ming-Hsuan Yang, and Sergey Tulyakov.
\newblock {InfiniCity}: Infinite-scale city synthesis.
\newblock In \emph{Proceedings of the IEEE/CVF International Conference on Computer Vision}, pages 22808--22818, 2023.

\bibitem[Ling et~al.(2024)Ling, Sheng, Tu, Zhao, Xin, Wan, Yu, Guo, Yu, Lu, Li, Sun, Ashok, Mukherjee, Kang, Kong, Hua, Zhang, Benes, and Bera]{Ling_2024_CVPR}
Lu Ling, Yichen Sheng, Zhi Tu, Wentian Zhao, Cheng Xin, Kun Wan, Lantao Yu, Qianyu Guo, Zixun Yu, Yawen Lu, Xuanmao Li, Xingpeng Sun, Rohan Ashok, Aniruddha Mukherjee, Hao Kang, Xiangrui Kong, Gang Hua, Tianyi Zhang, Bedrich Benes, and Aniket Bera.
\newblock Dl3dv-10k: A large-scale scene dataset for deep learning-based 3d vision.
\newblock In \emph{Proceedings of the IEEE/CVF Conference on Computer Vision and Pattern Recognition (CVPR)}, pages 22160--22169, 2024.

\bibitem[Lipman et~al.(2023)Lipman, Chen, Ben-Hamu, Nickel, and Le]{lipman2023flowmatching}
Yaron Lipman, Ricky T.~Q. Chen, Heli Ben-Hamu, Maximilian Nickel, and Matt Le.
\newblock Flow matching for generative modeling.
\newblock In \emph{International Conference on Learning Representations}, 2023.

\bibitem[Liu et~al.(2025)Liu, Tayal, Wang, Zarzar, Monnier, Tertikas, Duan, Toisoul, Zhang, Neverova, Vedaldi, Shapovalov, and Novotny]{liu2025uco3d}
Xingchen Liu, Piyush Tayal, Jianyuan Wang, Jesus Zarzar, Tom Monnier, Konstantinos Tertikas, Jiali Duan, Antoine Toisoul, Jason~Y. Zhang, Natalia Neverova, Andrea Vedaldi, Roman Shapovalov, and David Novotny.
\newblock {UnCommon Objects in 3D}, 2025.

\bibitem[M{\"u}ller et~al.(2006)M{\"u}ller, Wonka, Haegler, Ulmer, and Van~Gool]{muller2006procedural}
Pascal M{\"u}ller, Peter Wonka, Simon Haegler, Andreas Ulmer, and Luc Van~Gool.
\newblock Procedural modeling of buildings.
\newblock \emph{ACM Transactions on Graphics}, 25\penalty0 (3):\penalty0 614--623, 2006.

\bibitem[Nan et~al.(2025)Nan, Xie, Zhou, Fan, Yang, Chen, Li, Yang, and Tai]{nan2024openvid}
Kepan Nan, Rui Xie, Penghao Zhou, Tiehan Fan, Zhenheng Yang, Zhijie Chen, Xiang Li, Jian Yang, and Ying Tai.
\newblock {OpenVid-1M}: A large-scale high-quality dataset for text-to-video generation.
\newblock In \emph{International Conference on Learning Representations (ICLR)}, 2025.

\bibitem[Nauata et~al.(2020)Nauata, Chang, Cheng, Mori, and Furukawa]{nauata2020house}
Nelson Nauata, Kai-Hung Chang, Chin-Yi Cheng, Greg Mori, and Yasutaka Furukawa.
\newblock House-gan: Relational generative adversarial networks for graph-constrained house layout generation.
\newblock In \emph{Computer Vision--ECCV 2020: 16th European Conference, Glasgow, UK, August 23--28, 2020, Proceedings, Part I 16}, pages 162--177. Springer, 2020.

\bibitem[Paananen et~al.(2024)Paananen, Oppenlaender, and Visuri]{paananen2024ideation}
Ville Paananen, Jonas Oppenlaender, and Aku Visuri.
\newblock Using text-to-image generation for architectural design ideation.
\newblock \emph{International Journal of Architectural Computing}, 22\penalty0 (3):\penalty0 458--474, 2024.

\bibitem[Paulson(1976)]{paulson1976designing}
Boyd~C. Paulson, Jr.
\newblock Designing to reduce construction costs.
\newblock \emph{Journal of the Construction Division}, 102\penalty0 (4):\penalty0 587--592, 1976.

\bibitem[Rafailov et~al.(2023)Rafailov, Sharma, Mitchell, Ermon, Manning, and Finn]{rafailov2023direct}
Rafael Rafailov, Archit Sharma, Eric Mitchell, Stefano Ermon, Christopher~D. Manning, and Chelsea Finn.
\newblock Direct preference optimization: Your language model is secretly a reward model.
\newblock \emph{arXiv preprint arXiv:2305.18290}, 2023.

\bibitem[Ravi et~al.(2025)]{ravi2024sam2}
Nikhila Ravi et~al.
\newblock {SAM} 2: Segment anything in images and videos.
\newblock In \emph{International Conference on Learning Representations (ICLR)}, 2025.

\bibitem[Reizenstein et~al.(2021)Reizenstein, Shapovalov, Henzler, Sbordone, Labatut, and Novotny]{reizenstein2021co3d}
Jeremy Reizenstein, Roman Shapovalov, Philipp Henzler, Luca Sbordone, Patrick Labatut, and David Novotny.
\newblock Common objects in 3d: Large-scale learning and evaluation of real-life 3d category reconstruction.
\newblock In \emph{Proceedings of the IEEE/CVF International Conference on Computer Vision (ICCV)}, pages 10901--10911, 2021.

\bibitem[Ren et~al.(2025)Ren, Shen, Huang, Ling, Lu, Nimier-David, Müller, Keller, Fidler, and Gao]{ren2025gen3c}
Xuanchi Ren, Tianchang Shen, Jiahui Huang, Huan Ling, Yifan Lu, Merlin Nimier-David, Thomas Müller, Alexander Keller, Sanja Fidler, and Jun Gao.
\newblock Gen3c: 3d-informed world-consistent video generation with precise camera control.
\newblock In \emph{Proceedings of the IEEE/CVF Conference on Computer Vision and Pattern Recognition}, 2025.

\bibitem[Rombach et~al.(2022)Rombach, Blattmann, Lorenz, Esser, and Ommer]{rombach2022highresolution}
Robin Rombach, Andreas Blattmann, Dominik Lorenz, Patrick Esser, and Bj{\"o}rn Ommer.
\newblock High-resolution image synthesis with latent diffusion models, 2022.

\bibitem[Sch{\"o}nberger and Frahm(2016)]{schoenberger2016sfm}
Johannes~L. Sch{\"o}nberger and Jan-Michael Frahm.
\newblock Structure-from-motion revisited.
\newblock In \emph{Proceedings of the IEEE Conference on Computer Vision and Pattern Recognition (CVPR)}, pages 4104--4113, 2016.

\bibitem[Sch{\"o}nberger et~al.(2016)Sch{\"o}nberger, Zheng, Frahm, and Pollefeys]{schoenberger2016mvs}
Johannes~L. Sch{\"o}nberger, Enliang Zheng, Jan-Michael Frahm, and Marc Pollefeys.
\newblock Pixelwise view selection for unstructured multi-view stereo.
\newblock In \emph{Proceedings of the European Conference on Computer Vision (ECCV)}, pages 501--518, 2016.

\bibitem[Shabani et~al.(2023)Shabani, Hosseini, and Furukawa]{shabani2023housediffusion}
Mohammad~Amin Shabani, Sepidehsadat Hosseini, and Yasutaka Furukawa.
\newblock {HouseDiffusion}: Vector floorplan generation via a diffusion model with discrete and continuous denoising.
\newblock In \emph{Proceedings of the IEEE/CVF Conference on Computer Vision and Pattern Recognition}, pages 5466--5475, 2023.

\bibitem[Shen(2024)]{shen2024streetview360atoz}
Everett Shen.
\newblock {StreetView360AtoZ}: A 360-degree equirectangular street view dataset.
\newblock \url{https://huggingface.co/datasets/everettshen/StreetView360AtoZ}, 2024.
\newblock Hugging Face dataset.

\bibitem[Simonyan and Zisserman(2015)]{simonyan2015vgg}
Karen Simonyan and Andrew Zisserman.
\newblock Very deep convolutional networks for large-scale image recognition.
\newblock In \emph{International Conference on Learning Representations}, 2015.

\bibitem[Sun et~al.(2024)Sun, Chen, Liu, Chen, Duan, Zhang, and Wang]{sun2024dimensionx}
Wenqiang Sun, Shuo Chen, Fangfu Liu, Zilong Chen, Yueqi Duan, Jun Zhang, and Yikai Wang.
\newblock Dimensionx: Create any 3d and 4d scenes from a single image with controllable video diffusion.
\newblock \emph{arXiv preprint arXiv:2411.04928}, 2024.

\bibitem[Team(2024)]{qwen2.5}
Qwen Team.
\newblock Qwen2.5: A party of foundation models, 2024.

\bibitem[Team(2025)]{hunyuan3d22025tencent}
Tencent~Hunyuan3D Team.
\newblock Hunyuan3d 2.0: Scaling diffusion models for high resolution textured 3d assets generation, 2025.

\bibitem[Tsai and Hariharan(2025)]{Tsai20253D}
I-Ting Tsai and Bharath Hariharan.
\newblock {3D} synthesis for architectural design.
\newblock In \emph{WACV}, pages 4799--4809, 2025.

\bibitem[Tung et~al.(2024)Tung, Chou, Cai, Yang, Zhang, Wetzstein, Hariharan, and Snavely]{tung2024megascenes}
Joseph Tung, Gene Chou, Ruojin Cai, Guandao Yang, Kai Zhang, Gordon Wetzstein, Bharath Hariharan, and Noah Snavely.
\newblock Megascenes: Scene-level view synthesis at scale.
\newblock In \emph{ECCV}, 2024.

\bibitem[Voleti et~al.(2024)]{voleti2024sv3d}
Vikram Voleti et~al.
\newblock {SV3D}: Novel multi-view synthesis and {3D} generation from a single image using latent video diffusion.
\newblock In \emph{European Conference on Computer Vision (ECCV)}, 2024.

\bibitem[Wallace et~al.(2024)Wallace, Dang, Rafailov, Zhou, Lou, Purushwalkam, Ermon, Xiong, Joty, and Naik]{wallace2023diffusiondpo}
Bram Wallace, Meihua Dang, Rafael Rafailov, Linqi Zhou, Aaron Lou, Senthil Purushwalkam, Stefano Ermon, Caiming Xiong, Shafiq Joty, and Nikhil Naik.
\newblock Diffusion model alignment using direct preference optimization.
\newblock In \emph{Proceedings of the IEEE/CVF Conference on Computer Vision and Pattern Recognition}, 2024.

\bibitem[Wan et~al.(2025)Wan, Wang, Ai, Wen, Mao, Xie, Chen, Yu, Zhao, Yang, Zeng, Wang, Zhang, Zhou, Wang, Chen, Zhu, Zhao, Yan, Huang, Feng, Zhang, Li, Wu, Chu, Feng, Zhang, Sun, Fang, Wang, Gui, Weng, Shen, Lin, Wang, Wang, Zhou, Wang, Shen, Yu, Shi, Huang, Xu, Kou, Lv, Li, Liu, Wang, Zhang, Huang, Li, Wu, Liu, Pan, Zheng, Hong, Shi, Feng, Jiang, Han, Wu, and Liu]{wan2025}
Team Wan, Ang Wang, Baole Ai, Bin Wen, Chaojie Mao, Chen-Wei Xie, Di Chen, Feiwu Yu, Haiming Zhao, Jianxiao Yang, Jianyuan Zeng, Jiayu Wang, Jingfeng Zhang, Jingren Zhou, Jinkai Wang, Jixuan Chen, Kai Zhu, Kang Zhao, Keyu Yan, Lianghua Huang, Mengyang Feng, Ningyi Zhang, Pandeng Li, Pingyu Wu, Ruihang Chu, Ruili Feng, Shiwei Zhang, Siyang Sun, Tao Fang, Tianxing Wang, Tianyi Gui, Tingyu Weng, Tong Shen, Wei Lin, Wei Wang, Wei Wang, Wenmeng Zhou, Wente Wang, Wenting Shen, Wenyuan Yu, Xianzhong Shi, Xiaoming Huang, Xin Xu, Yan Kou, Yangyu Lv, Yifei Li, Yijing Liu, Yiming Wang, Yingya Zhang, Yitong Huang, Yong Li, You Wu, Yu Liu, Yulin Pan, Yun Zheng, Yuntao Hong, Yupeng Shi, Yutong Feng, Zeyinzi Jiang, Zhen Han, Zhi-Fan Wu, and Ziyu Liu.
\newblock Wan: Open and advanced large-scale video generative models.
\newblock \emph{arXiv preprint arXiv:2503.20314}, 2025.

\bibitem[Wang et~al.(2026)Wang, Zha, Cheng, Yang, Purkait, and Li]{wang2026orbital}
Rong Wang, Ruyi Zha, Ziang Cheng, Jiayu Yang, Pulak Purkait, and Hongdong Li.
\newblock Towards realistic and consistent orbital video generation via {3D} foundation priors.
\newblock \emph{arXiv preprint arXiv:2604.12309}, 2026.

\bibitem[Wang et~al.(2024)Wang, Yuan, Wang, Li, Chen, Xia, Luo, and Shan]{wang2024motionctrl}
Zhouxia Wang, Ziyang Yuan, Xintao Wang, Yaowei Li, Tianshui Chen, Menghan Xia, Ping Luo, and Ying Shan.
\newblock Motionctrl: A unified and flexible motion controller for video generation.
\newblock In \emph{ACM SIGGRAPH 2024 Conference Papers}, 2024.

\bibitem[Wibowo(2022)]{cadcrowd_animation_costs}
Mario Wibowo.
\newblock What are architectural animation costs, rates, \& {3D} design pricing for companies?
\newblock Cad Crowd blog, 2022.
\newblock Last updated 2022-08-05. Accessed 2026-11-12.

\bibitem[Xiang et~al.(2025)]{xiang2025trellis2}
Jianfeng Xiang et~al.
\newblock Native and compact structured latents for {3D} generation.
\newblock \emph{arXiv preprint arXiv:2512.14692}, 2025.

\bibitem[Xie et~al.(2024)Xie, Chen, Hong, and Liu]{xie2024citydreamer}
Haozhe Xie, Zhaoxi Chen, Fangzhou Hong, and Ziwei Liu.
\newblock {CityDreamer}: Compositional generative model of unbounded {3D} cities.
\newblock In \emph{Proceedings of the IEEE/CVF Conference on Computer Vision and Pattern Recognition}, pages 9666--9675, 2024.

\bibitem[Yang et~al.(2024)Yang, Chen, Pan, Yao, Chen, Ngo, and Mei]{yang2024hi3d}
Haibo Yang, Yang Chen, Yingwei Pan, Ting Yao, Zhineng Chen, Chong-Wah Ngo, and Tao Mei.
\newblock Hi3d: Pursuing high-resolution image-to-3d generation with video diffusion models.
\newblock In \emph{ACM MM}, 2024.

\bibitem[Yin et~al.(2026)Yin, Shi, Guo, and Wang]{yin2026vigor}
Tengjiao Yin, Jinglei Shi, Heng Guo, and Xi Wang.
\newblock {VIGOR}: {VIdeo Geometry-Oriented Reward} for temporal generative alignment, 2026.

\bibitem[Yu et~al.(2025)Yu, Hu, Xing, and Shan]{Yu_2025_ICCV}
Mark Yu, Wenbo Hu, Jinbo Xing, and Ying Shan.
\newblock Trajectorycrafter: Redirecting camera trajectory for monocular videos via diffusion models.
\newblock In \emph{Proceedings of the IEEE/CVF International Conference on Computer Vision (ICCV)}, pages 100--111, 2025.

\bibitem[Yu et~al.(2024)Yu, Xing, Yuan, Hu, Li, Huang, Gao, Wong, Shan, and Tian]{yu2024viewcrafter}
Wangbo Yu, Jinbo Xing, Li Yuan, Wenbo Hu, Xiaoyu Li, Zhipeng Huang, Xiangjun Gao, Tien-Tsin Wong, Ying Shan, and Yonghong Tian.
\newblock Viewcrafter: Taming video diffusion models for high-fidelity novel view synthesis.
\newblock \emph{arXiv preprint arXiv:2409.02048}, 2024.

\bibitem[Zhang et~al.(2023)Zhang, Rao, and Agrawala]{zhang2023adding}
Lvmin Zhang, Anyi Rao, and Maneesh Agrawala.
\newblock Adding conditional control to text-to-image diffusion models, 2023.

\bibitem[Zhang et~al.(2018)Zhang, Isola, Efros, Shechtman, and Wang]{zhang2018lpips}
Richard Zhang, Phillip Isola, Alexei~A. Efros, Eli Shechtman, and Oliver Wang.
\newblock The unreasonable effectiveness of deep features as a perceptual metric.
\newblock In \emph{Proceedings of the IEEE Conference on Computer Vision and Pattern Recognition}, pages 586--595, 2018.

\bibitem[Zhao et~al.(2025)Zhao, Lin, Lin, Yan, Li, Yang, Wang, Lee, and Wang]{zhao2024genxd}
Yuyang Zhao, Chung-Ching Lin, Kevin Lin, Zhiwen Yan, Linjie Li, Zhengyuan Yang, Jianfeng Wang, Gim~Hee Lee, and Lijuan Wang.
\newblock Genxd: Generating any 3d and 4d scenes.
\newblock In \emph{ICLR}, 2025.

\bibitem[Zhou et~al.(2025)Zhou, Gao, Voleti, Vasishta, Yao, Boss, Torr, Rupprecht, and Jampani]{zhou2025stablevirtualcamera}
Jensen~(Jinghao) Zhou, Hang Gao, Vikram Voleti, Aaryaman Vasishta, Chun-Han Yao, Mark Boss, Philip Torr, Christian Rupprecht, and Varun Jampani.
\newblock Stable virtual camera: Generative view synthesis with diffusion models.
\newblock \emph{arXiv preprint arXiv:2503.14489}, 2025.

\bibitem[Zhou et~al.(2018)Zhou, Tucker, Flynn, Fyffe, and Snavely]{zhou2018stereo}
Tinghui Zhou, Richard Tucker, John Flynn, Graham Fyffe, and Noah Snavely.
\newblock Stereo magnification: Learning view synthesis using multiplane images.
\newblock \emph{ACM Transactions on Graphics (TOG)}, 37\penalty0 (4):\penalty0 1--12, 2018.

\end{thebibliography}
}

\clearpage
\appendix
\clearpage
\maketitlesupplementary

Our project website presents video demonstrations of our results,
including comparisons across all methods on shared inputs,
alternative design completions, intermediate training stages,
and ablation studies. All videos are available at
\url{https://itingtsai.github.io/DesignOrbit/}.

\section{Additional Ablation Studies}
\label{sec:add_ablation}

\begin{table*}[t]
  \centering
\caption{\textbf{Ablations of data construction and training.}
Results on architectural inputs, grouped by experimental question. 
Full measures orbit completeness~\req{4} and, through its requirement that all frames register, consistency~\req{3}; tags in each group header name the requirements its question concerns.
Under \emph{Data Verifier}, Reg.$=81$ requires all frames to register; the column labeled $\tau\geq351^\circ$ indicates coverage filtering using
$C(x)\geq\tau$ with $\tau=351^\circ$.
(\xmark{}: check removed; --: no verifier; rank: coverage-ranked DPO pairs.)
Wan2.1 and HunyuanVideo are evaluated at $832\times480$ and $608\times352$, respectively ($W\times H$), with 81 frames at 15 fps.
$^{\dagger}$Rows with \cmark{}\ in both columns use the full training-data
verifier $V$, which additionally applies the directional-agreement
($A(x) \geq 0.90$) and radius-consistency ($\mathrm{CV}_r(x) \leq 0.30$) checks (Sec.~\ref{sec:camera_coverage}).
}
  \label{tab:ablation2}
  \footnotesize
  \setlength{\tabcolsep}{4pt}
  \begin{tabularx}{\textwidth}{@{}X cc cc@{}}
    \toprule
    \corner{WanVideo $832{\times}480$ (W$\times$H), HunyuanVideo $608{\times}352$ (W$\times$H), 15 fps, 81 frames} & \multicolumn{2}{c}{Data Verifier$^{\dagger}$} & \multicolumn{2}{c}{Model Inference} \\
    \cmidrule(lr){2-3} \cmidrule(lr){4-5}
    Variant & Reg.$=81$ & $\tau$ $\geq351^\circ$ & 
    Full $\uparrow$
    & 
    Cov.
    $\uparrow$ \\
    \midrule
    \multicolumn{5}{@{}l}{\textit{(a) Can existing datasets replace the procedural buildings? (\req{1}, \req{2}, \req{4})}} \\

    \textbf{Procedural buildings, inpainted background (Stage~I)} & \cmark & \cmark & \textbf{11.70\%} & \textbf{301.15$^\circ$} \\

    Objaverse, 3D buildings, inpainted background & \cmark & \cmark & 1.60\% & 271.61$^\circ$ \\

    \rowcolor{gray!10}\multicolumn{5}{@{}l}{\textcolor{gray}{\textit{For the inpainted Objaverse variant, same Stage~I recipe and verifier.}}} \\
    \rowcolor{gray!10}\multicolumn{5}{@{}l}{\textcolor{gray}{\textit{A third to half of its architecture-category assets are not buildings. We hand-picked the subset; procedural buildings need none.}}} \\

    Objaverse, 3D buildings, raw renders, no inpainting, no verifier & -- & -- & 0.00\% & 145.08$^\circ$ \\

    uCO3D, real object videos & -- & -- & 
    1.30\% 
    & 174.42$^\circ$ \\

    DL3DV-10K, real scene videos & -- & -- & 3.70\% & 189.05$^\circ$ \\
    \rowcolor{gray!10}\multicolumn{5}{@{}l}{\textcolor{gray}{\textit{Real captures, used as released without the verifier: uCO3D circles objects, DL3DV-10K captures scenes; both vary in viewpoint height and distance.}}} \\
    \rowcolor{gray!10}\multicolumn{5}{@{}l}{\textcolor{gray}{\textit{These datasets do not guarantee closed, evenly paced architectural orbits.
    We test whether real but imperfect orbits can replace our composites.}}} \\

    DL3DV + OpenVid + RealEstate10K + Panda-70M, drone orbit videos only & -- & -- & 5.90\% & 184.58$^\circ$ \\
    \rowcolor{gray!10}\multicolumn{5}{@{}l}{\textcolor{gray}{\textit{Drone footage often contains noisy camera motion and lacks clean, closed-loop orbits. We manually curate training samples from multiple datasets.}}} \\
    
    \midrule
    \multicolumn{5}{@{}l}{\textit{(b) Does Stage~I need loop-closed inpainting \emph{and} last-frame conditioning? (\req{1}, \req{4})}} \\

    \textbf{Loop-closed inpainting + verifier (FLF2V) (Stage~I)} & \cmark & \cmark & \textbf{11.70\%} & \textbf{301.15$^\circ$} \\

    Single-pass inpainting: the loop breaks & \cmark & \cmark & 0.00\% & $95.19^\circ$ \\

    Without last-frame conditioning (I2V) & \cmark & \cmark & 9.60\% & $298.01^\circ$ \\

    \rowcolor{gray!10}\multicolumn{5}{@{}l}{\textcolor{gray}{\textit{
    Without last-frame conditioning, the generated ending can exhibit an abrupt brightness or sharpness change.
    }}} \\

    With real panorama background, StreetView360AtoZ (FLF2V) & -- & -- & 7.80\% & $265.80^\circ$ \\

    Raw renders, no inpainting, no verifier: Naive & -- & -- & 0.00\% & $70.74^\circ$ \\
    \rowcolor{gray!10}\multicolumn{5}{@{}l}{\textcolor{gray}{\textit{Panorama and raw-render clips are not inpainted; their orbits close by construction, so no verifier is applied.}}} \\

    Wan2.1-FLF2V, Base Model ($832\times480$) & -- & -- & 0.00\% & $26.36^\circ$ \\

    \midrule
    \multicolumn{5}{@{}l}
    {\textit{(c) Does the verifier need both the registration and coverage check? (\req{3}, \req{4})}} \\
    \textbf{Round 2, registration + coverage filters (Stage~II)} & \cmark & \cmark & \textbf{16.50\%} & \textbf{318.21$^\circ$} \\
    Round 2, registration only & \cmark & \xmark & 3.70\% & $288.93^\circ$ \\
    Round 2, no filters & \xmark & \xmark & 0.00\% & $250.86^\circ$ \\
    \textbf{Round 1, registration + coverage filters (Stage~I)} & \cmark & \cmark & \textbf{11.70\%} & \textbf{301.15$^\circ$} \\
    Round 1, registration only & \cmark & \xmark & 0.00\% & $223.77^\circ$ \\
    Round 1, no filters & \xmark & \xmark & 1.10\% & $251.22^\circ$ \\
    \midrule
    \multicolumn{5}{@{}l}{\textit{(d) Are the synthetic composites still needed after round 1? (\req{4})}} \\

    \textbf{Round 2, synthetic clips + own verified generations (Stage~II)} & \cmark & \cmark & \textbf{16.50\%} & \textbf{318.21$^\circ$} \\

    Round 2, own verified generations only & \cmark & \cmark & 11.20\% & $313.72^\circ$ \\

    \midrule
    \multicolumn{5}{@{}l}
    {\textit{(e) At rounds 2 and 3, SFT with two-filter verification or coverage-ranked DPO? (\req{4})}} \\
    \textbf{Round 3, coverage-ranked DPO (Stage~III-DPO)} & -- & rank & \textbf{92.60\%} & \textbf{349.04$^\circ$} \\
    Round 3, SFT with two-filter verification (Stage~III-SFT) & \cmark & \cmark & 40.40\% & $332.38^\circ$ \\
    Round 2, coverage-ranked DPO (Stage~II-DPO) & -- & rank & 81.40\% & $ 343.88^\circ$ \\
    Round 2, SFT with two-filter verification (Stage~II) & \cmark & \cmark & 16.50\% & $318.21^\circ$ \\

    \midrule
    \multicolumn{5}{@{}l}{
      \textit{(f) Does the training recipe generalize to another base model? (\req{4})}
    } \\

    \textbf{HunyuanVideo, Stage~III-DPO}
      & -- & rank
      & \textbf{38.80\%} & \textbf{325.72$^\circ$} \\

    \rowcolor{gray!10}\multicolumn{5}{@{}l}{\textcolor{gray}{\textit{HunyuanVideo-I2V has no last-frame input, so we add one through zero-initialized input channels learned only from our orbit clips.}}} \\
    \rowcolor{gray!10}\multicolumn{5}{@{}l}{\textcolor{gray}{\textit{It also runs at a lower resolution. The recipe still transfers: each round raises coverage, from $52.01^\circ$ to $325.72^\circ$ after DPO.}}} \\

    HunyuanVideo, Stage~II
      & \cmark & \cmark
      & 4.80\% & 291.53$^\circ$ \\

    HunyuanVideo, Stage~I
      & \cmark & \cmark
      & 
      3.40\% 
      & 268.23$^\circ$ \\

    HunyuanVideo-I2V, Base Model ($608\times352$)
      & -- & --
      & 0.00\% & 52.01$^\circ$ \\

    \bottomrule
  \end{tabularx}
\end{table*}

\subsection{Data Construction and Training Pipeline}
\label{sec:training_ablation}

\begin{figure*}[t]
  \centering
  \includegraphics[width=\textwidth]{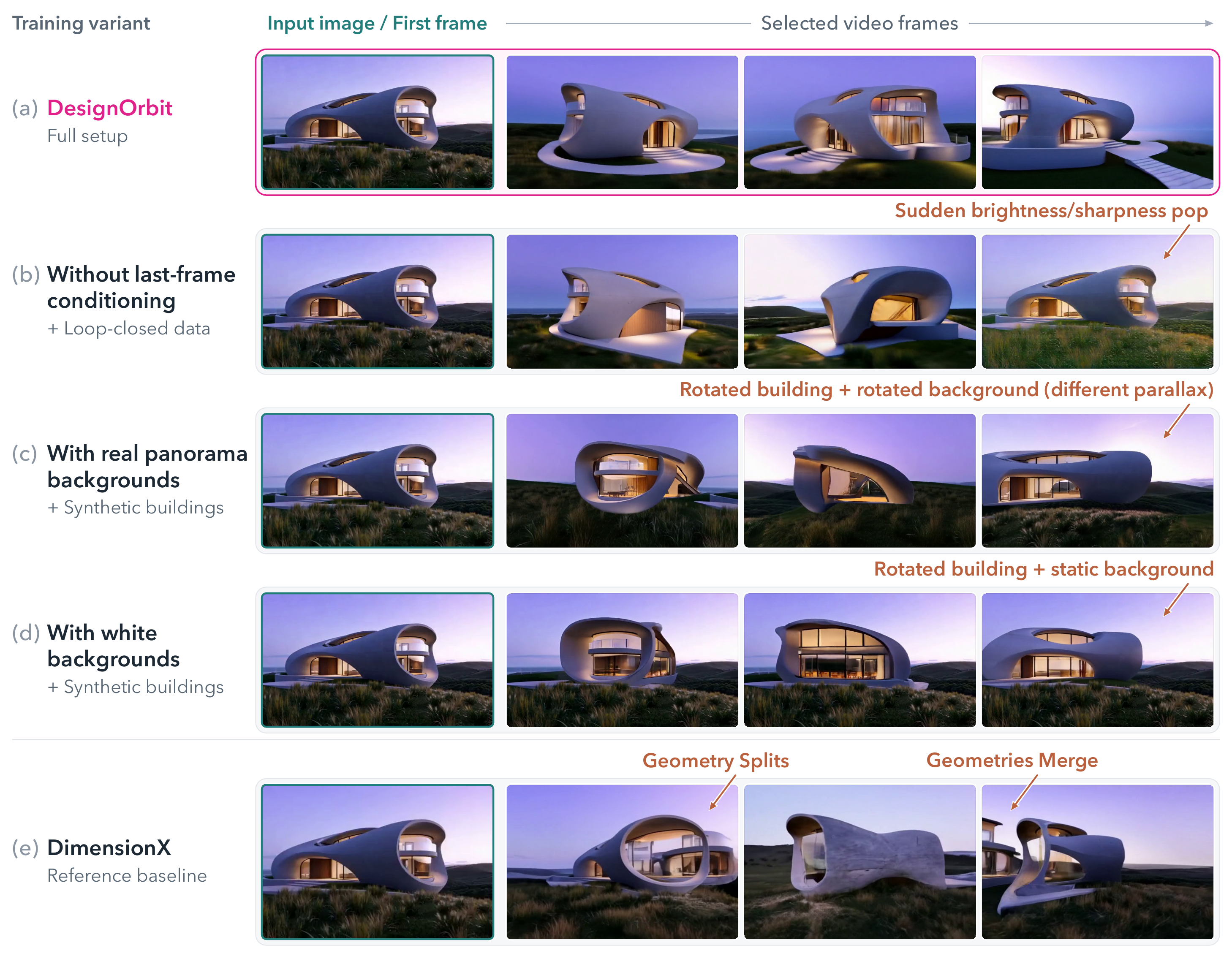}
  \caption{\textbf{Ablation of training data and conditioning.}
  All rows start from the same input image (left) and show selected frames from one generated orbit.
  (a)~The full setup orbits the building and its site coherently.
  (b)~Training on the same loop-closed data but without last-frame conditioning in the I2V model ends the orbit with a sudden brightness and sharpness pop.
  (c)~Replacing the inpainted surroundings with real panorama backgrounds teaches the model to rotate the building and background independently with different parallax.
  (d)~White backgrounds go further: only the building rotates, as on a turntable, while foreground and background stay fixed.
  (e)~DimensionX, shown for reference, splits and merges the building during the orbit. Videos of these and further ablations are on the supplementary web page.}
  \label{fig:ablation}
\end{figure*}

Table~\ref{tab:ablation2} evaluates our data construction and training choices on architectural inputs. As in main-paper \cref{sec:ablation}, each
variant is scored on orbit completeness~\req{4} with the Full and Cov.\ metrics of main-paper \cref{sec:quantitative}; Full also
requires all 81 frames to register, which checks consistency~\req{3}.
Context~\req{1} and plausibility~\req{2} cannot be measured by reconstruction, and we describe their effects qualitatively
(Fig.~\ref{fig:ablation}). Full uses only the registration and coverage checks; the training-data verifier $V$ additionally requires
$A(x)\geq0.90$ and $\mathrm{CV}_r(x)\leq0.30$
(Sec.~\ref{sec:camera_coverage}).

\paragraph{Data scarcity.}
Suitable training data for complete $360^\circ$ orbits around buildings in context are limited. 
uCO3D~\citep{liu2025uco3d} provides full-orbit videos, but of everyday objects rather than buildings in their sites. It therefore offers no architectural examples from which to learn plausibility~\req{2} and no architectural context~\req{1}.
MegaDepth~\cite{li2018megadepth} and
MegaScenes~\citep{tung2024megascenes} provide scene
observations from Internet photo collections rather than continuous orbit videos~(\req{4}). DL3DV-10K~\citep{Ling_2024_CVPR} includes real scene videos with varying camera heights and partial or full circular coverage,
rather than a fixed orbit protocol centered on individual
buildings~(\req{4}). Drone footage can also capture large urban or natural scenes from elevated viewpoints, without necessarily providing a complete
orbit around a single building~(\req{4}). Where a building is only partly circled, its rear is never observed, so these sources offer no supervision for completing the unseen sides~(\req{2}).
Objaverse~\citep{objaverse,objaverseXL} provides diverse 3D assets, but building models are not accompanied by their surroundings~(\req{1}). Each source therefore lacks context~\req{1}, supervision for architectural
plausibility~\req{2}, a closed orbit~\req{4}, or several of these, which motivates our construction of contextual architectural orbit data. We compare alternative training
data sources in our ablation study (\cref{tab:ablation2}).

\paragraph{(a) Alternative training data (\req{1}, \req{2}, \req{4}).}
Our Stage~I model reaches $301.15^\circ$ mean coverage and
$11.70\%$ full-orbit rate, above every training-data alternative tested.
The closest in mean coverage is Objaverse buildings processed with
the same inpainting and verifier, at $271.61^\circ$ and $1.60\%$. Both training sets contain the same number of clips and are evaluated
on the same test images. 
We hypothesize that the remaining gap partly reflects variation in the scale, pivot, and centering of Objaverse assets, which may produce less uniform framing in the rendered orbits. Procedural generation allows us to control these properties explicitly. The same assets as raw renders reach $145.08^\circ$ and $0.00\%$.
Thus, our pipeline improves mean coverage on Objaverse by
$126.53^\circ$.

Real captures used as released achieve lower coverage: DL3DV-10K~\citep{Ling_2024_CVPR} reaches $189.05^\circ$ and $3.70\%$ full-orbit rate, and uCO3D~\citep{liu2025uco3d} reaches $174.42^\circ$ and $1.30\%$. Much of the drone footage we inspected either lacks orbit motion around individual buildings or stops before completing a closed orbit.
We therefore manually curate clips showing orbit motion from DL3DV-10K~\citep{Ling_2024_CVPR}, OpenVid~\citep{nan2024openvid}, RealEstate10K~\citep{zhou2018stereo}, and Panda-70M~\citep{chen2024panda}.
Even with this curation, training on these clips reaches only $184.58^\circ$ mean coverage and $5.90\%$ full-orbit rate. 

Real captures have surroundings but rarely clean closed orbits around a building: scene videos seldom close the loop, and uCO3D~\citep{liu2025uco3d} circles objects with a viewpoint that varies in height and distance~(\req{4}). Objaverse~\citep{objaverse,objaverseXL} supplies buildings that need our inpainting to gain a site~(\req{1}); procedural buildings with inpainted surroundings supply both, and their rendered side and rear views also supervise plausibility~(\req{2}), which uCO3D's objects cannot.
These results support our combination of procedural buildings, contextual inpainting, and verification over the tested alternatives.
The pipeline also benefits existing Objaverse assets, but procedural
generation achieves higher coverage and full-orbit rate without requiring manual asset selection.

\paragraph{(b) Loop-closed data and endpoint conditioning (\req{1}, \req{4}).}
Split-and-condition inpainting provides more effective orbit supervision than single-pass inpainting, increasing mean coverage from $95.19^\circ$
to $301.15^\circ$ and full-orbit rate from $0.00\%$ to $11.70\%$.
Training on raw procedural renders, whose orbits are complete and consistent by construction and so need no verification, reaches only $70.74^\circ$ mean coverage.
Fig.~\ref{fig:ablation}(d) shows why: the resulting model rotates only the building, as on a turntable, while the foreground and background stay fixed, so the site shows no parallax~(\req{1}). This is the static-background behavior noted in main-paper \cref{sec:synthetic} and quantified in main-paper \cref{sec:ablation}.

Last-frame conditioning provides a smaller additional gain over I2V, increasing mean coverage from $298.01^\circ$ to $301.15^\circ$ and full-orbit rate from $9.60\%$ to $11.70\%$, and removes the brightness
and sharpness change where the orbit returns to its start~(\req{4}; Fig.~\ref{fig:ablation}b).
For the panorama variant, we sweep each equirectangular panorama from StreetView360AtoZ~\cite{shen2024streetview360atoz} through $360^\circ$ to form an 81-frame background clip and composite the rendered building over it, rotating both in sync. This variant reaches $265.80^\circ$ mean coverage and $7.80\%$ full-orbit rate. However, these training clips lack correct background parallax~(\req{1}).
In the generated videos, the building and background appear to rotate independently, so the frames no longer show one 3D scene~(\req{3}). Figure~\ref{fig:ablation} illustrates the effects of training data and endpoint conditioning.

\paragraph{(c) Registration and coverage checks (\req{3}, \req{4}).}
Adding the coverage check improves over registration filtering alone
in both training rounds. At round~2, full-orbit rate rises from $3.70\%$ to
$16.50\%$, and mean coverage increases from $288.93^\circ$ to
$318.21^\circ$. At round~1, registration-only filtering even
underperforms unfiltered training, while combining registration and
coverage gives the best results. 
Registration alone checks consistency~\req{3} but does not select complete orbits~\req{4}; the coverage check is needed for that.

\paragraph{(d) Retaining synthetic supervision (\req{4}).}
Synthetic clips remain beneficial after initialization. At round~2,
combining them with verified self-generated videos increases full-orbit rate from $11.20\%$ to $16.50\%$ and mean coverage from $313.72^\circ$
to $318.21^\circ$, compared with using verified generations alone.
We hypothesize that this is because the synthetic clips are the only training clips whose loop closure~\req{4} holds by construction rather than by verification.

\paragraph{(e) SFT with two-filter verification vs.\ coverage-ranked DPO (\req{4}).}
Coverage-ranked DPO outperforms SFT with two-filter verification
(registration and coverage, with the two trajectory checks of
Sec.~\ref{sec:camera_coverage}) at both rounds.
At round~2, DPO reaches an $81.40\%$ full-orbit rate and $343.88^\circ$ mean coverage, compared with $16.50\%$ and $318.21^\circ$ for SFT; at round~3, it reaches $92.60\%$ and $349.04^\circ$, compared with $40.40\%$ and $332.38^\circ$. SFT learns only from videos that pass
the verifier, whereas DPO also uses incomplete orbits as negative examples (main-paper \cref{sec:dpo}). Ranking by coverage therefore enforces
completeness~\req{4} more effectively than filtering on it, and participants still rate the unseen sides of the final model as compatible and usable~\req{2} (main-paper Tab.~\ref{tab:user_study}).

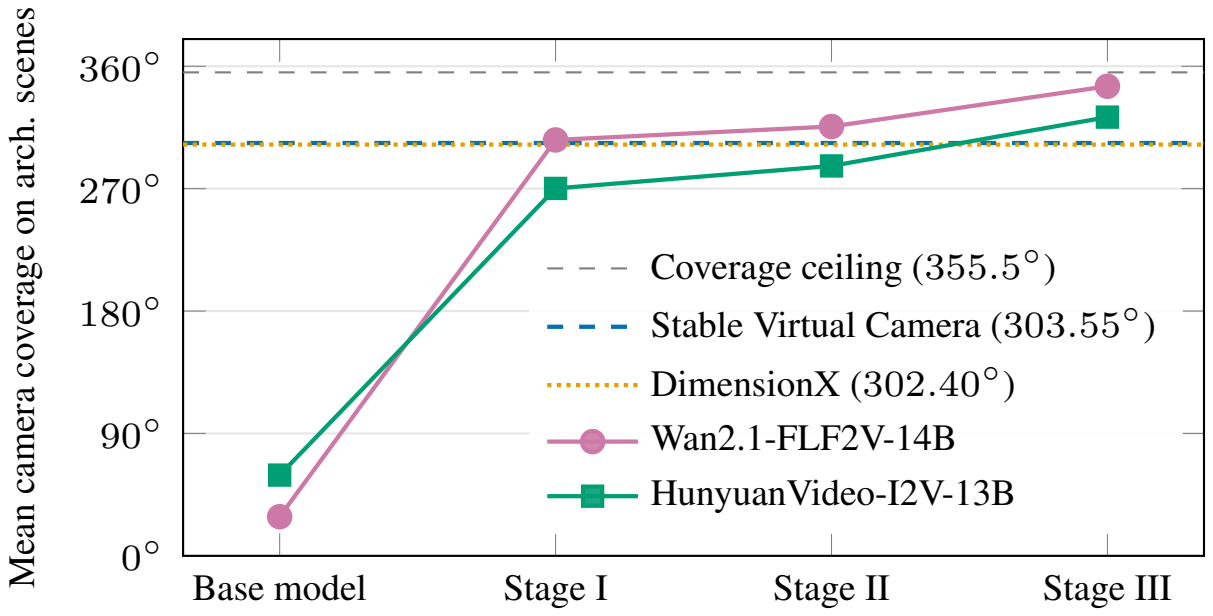
\begin{figure}[t]
  \centering
  \begin{tikzpicture}
  \definecolor{roundssvc}{HTML}{0072B2}
  \definecolor{roundshunyuan}{HTML}{009E73}
    \definecolor{coldimx}{HTML}{E69F00}  %
    \definecolor{mag}{HTML}{CC79A7}      %
  \begin{axis}[
    width=\linewidth, height=5.0cm,
    xmin=-0.35, xmax=3.35, ymin=0, ymax=380,
    xtick={0,1,2,3},
    xticklabels={
      {Base model},
      {Stage I},
      {Stage II},
      {Stage III}
    },
    xticklabel style={align=center, font=\scriptsize},
    ytick={0,90,180,270,360},
    yticklabel={$\pgfmathprintnumber{\tick}^\circ$},
    yticklabel style={font=\scriptsize},
    ylabel={Mean camera coverage on arch. scenes},
    ylabel style={font=\scriptsize},
    ymajorgrids=true,
    grid style={gray!20},
    legend style={
      font=\scriptsize,
      at={(0.98,0.04)},
      anchor=south east,
      draw=none,
      fill=white,
      fill opacity=0.9,
      text opacity=1
    },
    legend cell align=left,
  ]

  \addplot[
    domain=-0.35:3.35, samples=2, dashed, gray
  ] {355.5};
  \addlegendentry{Coverage ceiling ($355.5^\circ$)}

  \addplot[
    domain=-0.35:3.35, samples=2,
    dashed, thick, roundssvc
  ] {303.55};
  \addlegendentry{Stable Virtual Camera ($303.55^\circ$)}
  
  \addplot[
    domain=-0.35:3.35, samples=2,
    densely dotted, thick, coldimx
  ] {302.40};
  \addlegendentry{DimensionX ($302.40^\circ$)}

  \addplot[
    thick, mag, mark=*, mark size=2pt
  ] coordinates {
    (0,28.77)
    (1,305.81)
    (2,315.70)
    (3,345.29)
  };
  \addlegendentry{Wan2.1-FLF2V-14B}

  \addplot[
    thick, solid, roundshunyuan,
    mark=square*, mark size=1.8pt
  ] coordinates {
    (0,59.38)
    (1,270.16)
    (2,286.70)
    (3,322.59)
  };
  \addlegendentry{HunyuanVideo-I2V-13B}
  
  \end{axis}
  \end{tikzpicture}
 \caption{\textbf{Mean camera coverage increases with each training round.}
  Stage~I trains on synthetic orbit videos; Stage~II combines synthetic clips with verified model generations. Stage~III uses coverage-ranked preference pairs for DPO. All videos are evaluated at $480\times320$ ($W\times H$).}
  \label{fig:rounds}
\end{figure}

\paragraph{(f) Transfer to another backbone (\req{4}).}
The training recipe also improves HunyuanVideo, which we equip with
last-frame conditioning. At its native $608\times352$ resolution,
mean coverage increases from $52.01^\circ$ for the base model to
$268.23^\circ$, $291.53^\circ$, and $325.72^\circ$ across
Stages~I--III, reaching a 38.80\% full-orbit rate (\cref{tab:ablation2}).
At the $480\times320$ resolution, mean coverage increases from $59.38^\circ$ to $270.16^\circ$, $286.70^\circ$, and $322.59^\circ$ (\cref{fig:rounds}),
the same progressive improvement as for Wan2.1, demonstrating that the
training recipe transfers beyond Wan2.1.

\begin{table*}[t]
  \centering
\caption{\textbf{Coverage-only ranking with pure DPO gives the best reported coverage results.} We ablate the criteria used to construct Stage~III-DPO preference pairs and the auxiliary SFT weight in $\mathcal L=\mathcal L_{\mathrm{DPO}}+\lambda\mathcal L_{\mathrm{SFT}}$. Under \emph{Data Verifier}, a \cmark{} in All Frame Reg.\ requires all 81 frames of the winning video to register in COLMAP; ``rank'' denotes a ranking criterion, and \xmark{} denotes an unused criterion. 3D pts denotes the reconstructed point count. Multiple ranking criteria are combined as a weighted sum on a common degree scale. 
Under Model inference, all metrics are computed over the same $\sim$190
generated architecture videos for every variant; All Frame Reg.\ is the
percentage with all 81 frames registered.
}
  \label{tab:dpo_ablation}
  \footnotesize
  \setlength{\tabcolsep}{4pt}
  \begin{tabularx}{\textwidth}{@{}X ccc ccc@{}}
    \toprule
    & \multicolumn{3}{c}{Data Verifier} & \multicolumn{3}{c}{Model Inference} \\
    \cmidrule(lr){2-4} \cmidrule(lr){5-7}
    Variant ($\mathcal{L}=\mathcal{L}_{\mathrm{DPO}}+\lambda\mathcal{L}_{\mathrm{SFT}}$) & All Frame Reg. & $C(x)$$^\circ$ & 3D pts & All Frame Reg.$\uparrow$ & 
    Full $\uparrow$
    & 
    Cov.
    $\uparrow$ \\
    \midrule
    \multicolumn{7}{@{}l}{where $\lambda=0.0$}\\

    \textbf{StageIII-DPO} & \xmark & rank & \xmark & 
    \textbf{92.60\%}
    & \textbf{92.60\%} & \textbf{349.04$^\circ$} \\ 
    
    StageIII-DPO & \cmark & rank & \xmark &
    79.80\%
    & 77.70\% & 338.99$^\circ$ \\

    StageIII-DPO & \cmark & rank & rank & 
    89.90\%
    & 81.90\% & 346.30$^\circ$ \\

    StageIII-DPO & \xmark & rank & rank &
    91.50\%
    & 81.40\% & 346.30$^\circ$ \\

    StageIII-DPO & rank & rank & rank & 
    76.60\%
    & 73.90\% & 333.48$^\circ$ \\

    StageIII-DPO & rank & rank & \xmark & 
    83.50\%
    & 81.90\% & 346.39$^\circ$ \\
    
    \midrule
    \multicolumn{7}{@{}l}{where $\lambda=0.1$}\\

    StageIII-DPO & \xmark & rank & \xmark & 
    91.00\%
    & 88.80\% & 348.78$^\circ$ \\

    StageIII-DPO & \cmark & rank & \xmark &
    83.50\%
    & 75.00\% & 339.30$^\circ$ \\

    StageIII-DPO & \cmark & rank & rank & 
    92.00\%
    & 86.70\% & 347.24$^\circ$ \\
    
    StageIII-DPO & \xmark & rank & rank &
    89.90\%
    & 79.30\% & 343.18$^\circ$ \\

    StageIII-DPO & rank & rank & rank & 
    72.90\%
    & 72.30\% & 334.61$^\circ$ \\

    StageIII-DPO & rank & rank & \xmark &
    87.20\%
    & 80.90\% & 344.11$^\circ$ \\

    \bottomrule
  \end{tabularx}
\end{table*}

\subsection{DPO Ranking Criteria and Regularization}
\label{sec:dpo_ablation}

\paragraph{Preference ranking criteria.}
For each input image and prompt, we rank eight sampled videos and pair the highest-ranked video (winner) with the lowest-ranked video (loser), retaining the pair only if the winner's coverage score exceeds the loser's by at least $30^\circ$. As in the main method, the coverage score is
$S(x)=C(x)$ when computable and $0^\circ$ otherwise. We compare
coverage-only ranking with variants that additionally use frame registration, reconstructed 3D point counts, or both
(Table~\ref{tab:dpo_ablation}). Registration and point count measure consistency~\req{3}, whereas coverage measures completeness~\req{4}, so this ablation asks whether adding consistency criteria to the ranking helps. Registration is used either as a hard requirement that all 81 frames of the winner register in COLMAP or as a ranking criterion; point
counts are used for ranking. When combining ranking criteria, we rescale the auxiliary criteria to the same numerical range as camera coverage and compute a weighted sum, assigning weights $0.7/0.3$ to coverage and one
auxiliary criterion, or $0.54/0.23/0.23$ to coverage, registration, and point count.

With pure DPO ($\lambda=0$), coverage-only ranking achieves a $92.60\%$ full-orbit rate and $349.04^\circ$ mean coverage; the highest corresponding values among the other pure-DPO variants are $81.90\%$ and $346.39^\circ$. Coverage-only ranking also achieves the highest full-registration rate, $92.60\%$, despite having no registration
criterion; adding a hard full-registration requirement lowers this rate to $79.80\%$. At $\lambda=0$, every variant that adds a consistency criterion has lower full-registration and full-orbit rates than coverage-only
ranking, so ranking by completeness~\req{4} alone does not cost consistency~\req{3}. Since about $90\%$ of the Stage~II candidates already register all 81 frames (Sec.~\ref{sec:sampling}), consistency criteria separate winners from losers little and mainly dilute the coverage ranking, which may instead reward less camera motion (main-paper
\cref{sec:dpo}).

\paragraph{Auxiliary SFT loss.}
We also evaluate an auxiliary SFT term with $\lambda=0.1$. For coverage-only ranking, the full-orbit rate decreases from $92.60\%$ to $88.80\%$ while mean
coverage changes slightly, from $349.04^\circ$ to $348.78^\circ$.
The SFT term improves some variants: combining a hard full-registration requirement with ranking by coverage and point count achieves an $86.70\%$ full-orbit rate and $347.24^\circ$ mean coverage.
However, none of the evaluated configurations outperforms
coverage-only ranking with pure DPO on either reported coverage
metric. Qualitative inspection revealed no clear visual-quality
difference between coverage-only models trained with and without
the auxiliary SFT term. We therefore rank preference pairs by camera coverage alone and omit the auxiliary SFT term in our final DPO stage.

\section{Extended Baseline Comparison}
\label{sec:supp_baseline_comparison}

\paragraph{Clarification of Tab.~\ref{tab:user_study}.}
Each entry in main-paper Tab.~\ref{tab:user_study} reports the percentage of ratings preferring \method{} / the baseline; remaining responses are ties, neither, or no answer. Single-video comparisons have 100 ratings per baseline per criterion, and video-pair comparisons have 50. The overall preference rate of 91.4\% pools individual ratings across comparisons, excluding ties, neither, and no-answer responses.

\begin{table}[t]
\centering
\caption{\textbf{User study by input category (50 participants): \method{} is preferred across all categories, with the largest margins on sketches.}
Criteria map to requirements as in main-paper \cref{sec:user_study}: Present.\ pref.~\req{1}, Shape consis.~\req{3}, Views compat.~\req{2}, Views visible~\req{4}.
Entries show \% choosing \method{} / baseline; remaining responses are ties, neither, or no answer.
Columns 1--3 compare single videos; 4--6 compare video pairs.
$n$: ratings per category (single/pair).}
\label{tab:user_study_cat}

\footnotesize
\setlength{\tabcolsep}{1.5pt}
\renewcommand{\arraystretch}{1.1}

\begin{tabular*}{\columnwidth}
{@{\extracolsep{\fill}}lcccccc@{}}
\toprule
& \multicolumn{3}{c}{Single video}
& \multicolumn{3}{c}{Video pairs} \\
\cmidrule(lr){2-4}
\cmidrule(l){5-7}

Input category
& \shortstack{Present.\\pref.}
& \shortstack{Shape\\consis.}
& \shortstack{Views\\compat.}
& \shortstack{Views\\visible}
& \shortstack{Unseen\\variety}
& \shortstack{Usable\\alter.} \\
\midrule

\shortstack[l]{Architecture\\{\tiny $n=150/50$}}
& \textbf{69}/9
& \textbf{62}/4
& \textbf{59}/9
& \textbf{42}/6
& \textbf{56}/10
& \textbf{48}/4 \\

\addlinespace[2pt]

\shortstack[l]{Other in-context\\{\tiny $n=300/100$}}
& \textbf{75}/5
& \textbf{60}/4
& \textbf{56}/5
& \textbf{36}/3
& \textbf{41}/17
& \textbf{56}/9 \\

\addlinespace[2pt]

\shortstack[l]{Arch. sketch\\{\tiny $n=50/50$}}
& \textbf{88}/2
& \textbf{84}/0
& \textbf{76}/6
& \textbf{90}/2
& \textbf{72}/8
& \textbf{78}/6 \\

\bottomrule
\end{tabular*}
\end{table}

\subsection{User Study by Input Category}
\label{sec:user_study_category}

We further analyze user preferences across three input
categories: architecture, other design-in-context, and
architectural sketches (Table~\ref{tab:user_study_cat}).
Although \method{} is trained only on architectural
scenes, participants prefer it over the baselines
across all six criteria in every category.

For architecture, \method{} is preferred for presentation in context~\req{1}
(69\% vs.\ 9\%), shape consistency~\req{3} (62\% vs.\ 4\%), and view compatibility~\req{2} (59\% vs.\ 9\%).
On other design-in-context inputs, \method{} is also preferred on all six criteria, showing that it generalizes beyond the training domain.
The largest margins occur on architectural sketches, where \method{} is preferred in 88\% vs.\ 2\% of presentation ratings~\req{1} and 90\% vs.\ 2\% of views-visible ratings~\req{4}.

Across video-set comparisons, participants consistently
prefer \method{} for revealing unseen views~\req{4}, generating
diverse completions, and providing usable design alternatives.
This preference holds across all input categories.

\subsection{Diversity of Design Alternatives}
Stage~III optimizes completeness~\req{4} only, so we check that it does not reduce the variety of alternative completions of the unseen sides that the task asks for (main-paper \cref{sec:background}). For each of 120 image--text inputs, we generate two videos using different random seeds, yielding 120 pairs per method, and measure the perceptual diversity of each
pair using DIV-LPIPS~\cite{ipoke} and DIV-VGG~\cite{cinn}.
For each video pair, we compute LPIPS distances~\cite{zhang2018lpips}
and VGG-16 feature distances~\cite{simonyan2015vgg}, respectively, between frames at matching time indices and average over frames. Higher values indicate greater perceptual variation between the generated videos.
Because frames are compared at matching time indices, these distances also include viewpoint differences: if two orbits of the same input progress at different speeds or stop early, their frames at the same index show different viewpoints. III-DPO's orbits usually complete the full circle (Table~\ref{tab:cross_model_compare2}), so its scores are less affected by viewpoint differences than those of methods with incomplete orbits, and its lead is unlikely to come from this effect.

Table~\ref{tab:diversity} shows that \method{} III-DPO achieves
the highest values on both metrics. Its DIV-LPIPS remains similar
to Stage~II ($0.403$ vs.\ $0.401$) and Stable Virtual Camera
($0.399$), while DIV-VGG increases from $6.172$ at Stage~II
to $7.682$ after DPO. DimensionX scores lower on both metrics.
Together with Table~\ref{tab:cross_model_compare2}, these results suggest that Stage~III raises completeness~\req{4} without reducing the variety of design alternatives.

\begin{table}
  \centering
  \caption{\textbf{Diversity of generated design alternatives.}
  For each method, we evaluate 120 architecture video pairs (240 generated videos). Each pair uses the same image--text input and two different random seeds. 
  DIV-LPIPS and DIV-VGG measure perceptual distances between corresponding frames using LPIPS and VGG-16 features, respectively. Higher values indicate greater diversity.}
  \footnotesize
  \setlength{\tabcolsep}{6pt}
  \renewcommand{\arraystretch}{1.15}
  \begin{tabular*}{\linewidth}{@{\extracolsep{\fill}}l c c @{}}
    \toprule
    Method (All $480\times 320$ ($W\times H$))
    & DIV-LPIPS $\uparrow$
    & DIV-VGG $\uparrow$ \\
    \midrule
    \textbf{\method{} III-DPO}
    & \textbf{0.403}
    & \textbf{7.682} \\

    \method{} II
    & 0.401
    & 6.172 \\

    \method{} I
    & 0.397
    & 6.294 \\
    
    Stable Virtual Camera
    & 0.399
    & 6.599 \\

    DimensionX
    & 0.289
    & 4.620 \\
    \bottomrule
  \end{tabular*}
  \label{tab:diversity}
\end{table}

\subsection{Orbit Completeness}
\label{sec:supp_orbit_complete}

\begin{figure*}
    \centering
\includegraphics[width=0.68\linewidth]{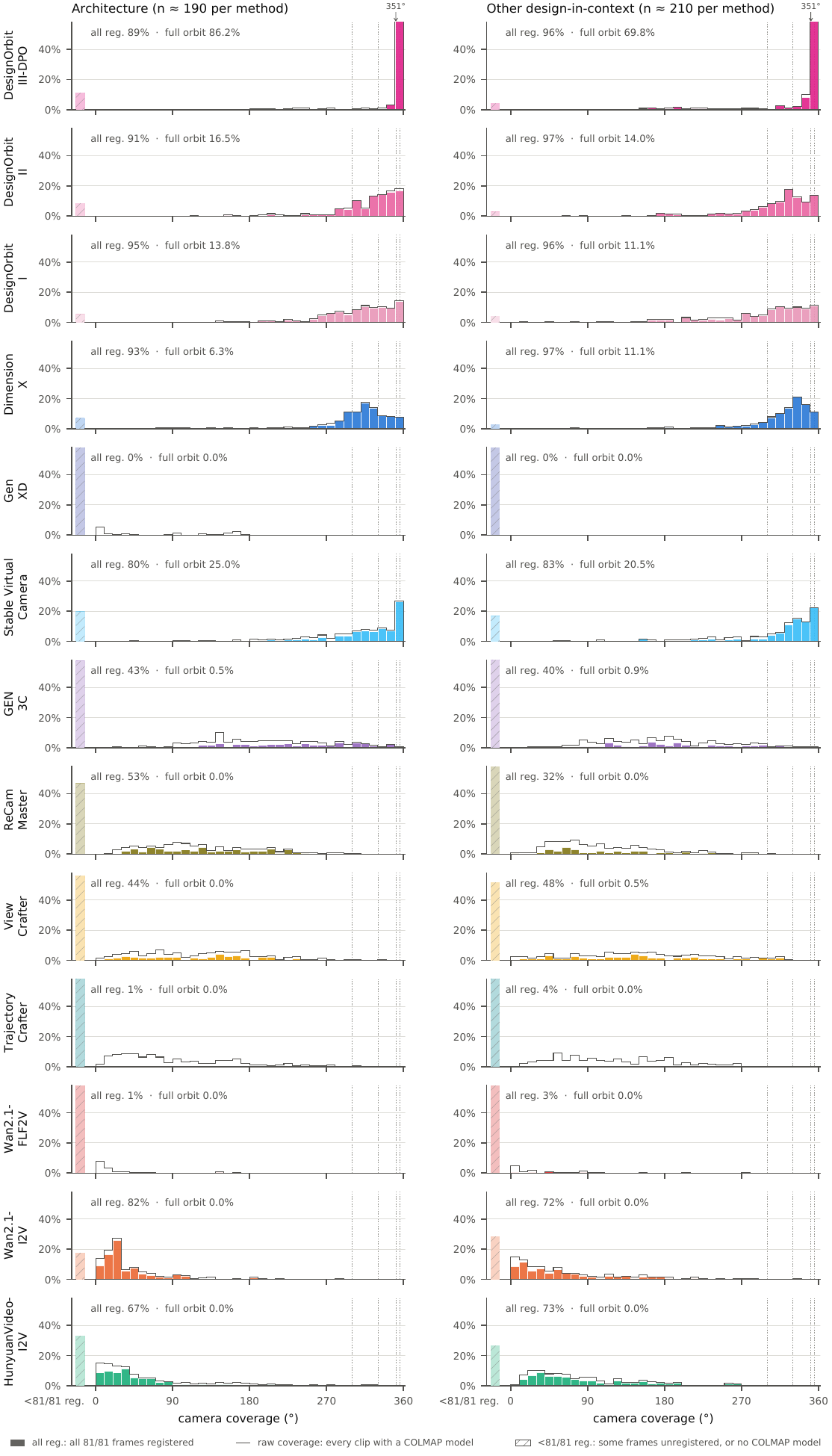}
\caption{\textbf{\method{} III-DPO achieves the highest full-orbit rates across both input categories.}
Coverage improves across training stages, while reconstruction-oriented models (DimensionX, GenXD) and camera-controlled baselines (Stable Virtual Camera, GEN3C, ReCamMaster, ViewCrafter, TrajectoryCrafter) rarely complete an orbit~(\req{4}). Base video models show little camera motion.
Annotations report the percentage of videos with all 81 frames registered
(all reg.) and the full-orbit rate (main-paper \cref{sec:quantitative}). Hatched bars indicate incomplete registration.
}

    \label{fig:hist_all}
\end{figure*}

Table~\ref{tab:cross_model_compare2} extends the main-paper comparison with intermediate training stages, base video models, and 3D generators. 
The table reports completeness~\req{4} per domain and consistency~\req{3} as foreground points; 
shaded rows describe what a designer sees in each output, including failures of context~\req{1} and plausibility~\req{2} that reconstruction cannot measure, tagged with the requirements each output misses.
A single failure can miss more than one requirement: a flattened or shape-changing design is inconsistent across views~\req{3} and also shows unseen sides that are not a plausible completion of the input~\req{2}; main-paper \cref{sec:quantitative} tags these failures by their primary requirement~\req{3}.
Figure~\ref{fig:hist_all} shows the coverage distributions and frame-registration outcomes for each method.

\paragraph{Effect of training stages (\req{4}).}
Each training stage improves orbit coverage in both domains.
On architecture, the full-orbit rate increases from $13.80\%$ at Stage~I to $16.50\%$ at Stage~II and $86.20\%$ after Stage~III-DPO (at $480\times320$).
The same progression holds on other design-in-context inputs:
$11.10\%$, $14.00\%$, and $69.80\%$, respectively.
Mean coverage also increases at each stage.
The largest improvement in near-complete orbit generation occurs
at the final DPO stage, including on inputs outside the training domain.

\paragraph{Base video models (\req{2}, \req{4}).}
None of the three base video models produces a full orbit in either domain.
On architecture, their mean coverage ranges from $28.77^\circ$
to $59.38^\circ$, compared with $305.81^\circ$ already achieved
by Stage~I.
Their outputs typically exhibit limited camera motion, panning,
or zooming, consistent with these low coverage values.
This comparison demonstrates the benefit of training specifically
for orbit generation.

\paragraph{3D generators (\req{1}, \req{2}).}
TRELLIS.2 and Hunyuan3D-2 support full camera orbits by rendering
a fixed 3D asset, so their orbit coverage is prescribed by construction.
However, the generated assets can be structurally incomplete and,
in our evaluation, do not include the surrounding scene.
They thus meet completeness~\req{4} by construction while missing
context~\req{1} and at times plausibility~\req{2}, so coverage alone cannot rank methods for ArchVis.

\begin{table*}[t]
  \centering
\caption{\textbf{\method{} achieves the highest full-orbit rates and reconstructs the most foreground points.}
We compare $\sim$190 architecture and $\sim$210 other design-in-context videos per method, normalized to 81 frames, 15\,fps, and $480\times320$ via aspect-ratio-preserving resizing and center cropping.
Full, Cov., and Pts.\ follow main-paper \cref{tab:combined_results}.
Shaded rows summarize qualitative observations, tagged with the
requirements each output misses;
3D generators have complete orbits~\req{4} by construction but lack surrounding context~\req{1}.}

  \footnotesize
  \setlength{\tabcolsep}{3pt}
  \renewcommand{\arraystretch}{1.1}
  \begin{tabular*}{\linewidth}{@{\extracolsep{\fill}}l cc cc c@{}}
    \toprule
    \corner{All video: aspect-ratio-preserving resized \& }
     & \multicolumn{4}{c}{Complete orbits?~\req{4} (ceiling for ideal sampling: $355.5^\circ$; $4.5^\circ$ per frame)} & \multicolumn{1}{c}{Consistent geometry?~\req{3}} \\
    \cmidrule(lr){2-5}\cmidrule(l){6-6}
    \corner{center-cropped to $480{\times}320$ (W$\times$H), 15\,fps, 81 frames} & \multicolumn{2}{c}{Architecture} & \multicolumn{2}{c}{Other design-in-context} & All categories\\
    \cmidrule(lr){2-3}\cmidrule(lr){4-5}\cmidrule(l){6-6}
    Method & 
    Full $\uparrow$
    & 
    Cov.
    $\uparrow$ &
    Full $\uparrow$
    & 
    Cov.
    $\uparrow$ & Pts. $\uparrow$ \\
    \midrule
    \textbf{\method{} III-DPO} & \textbf{86.20\%} & \textbf{345.29$^\circ$} & \textbf{69.80\%} & \textbf{332.35$^\circ$} & \textbf{1897} \\
    \sees{The design orbits in its context, with usable side and rear views; meets all four requirements.} \\
    \method{} II & 16.50\% & 315.70$^\circ$ & 14.00\% & 307.32$^\circ$ & -- \\
    \sees{The design orbits with less shape overfitting, but orbits are often incomplete~\req{4} and some geometry is inconsistent~\req{3}.} \\
    \method{} I & 13.80\% & 305.81$^\circ$ & 11.10\% & 291.40$^\circ$ & -- \\
    \sees{The design orbits, but orbits are often incomplete~\req{4}, some geometry is inconsistent~\req{3}, and shapes overfit to the cuboid training data~\req{2}.} \\
    \midrule
    DimensionX & 6.30\% & 302.40$^\circ$ & 11.10\% & 316.62$^\circ$ & 1792 \\
    \sees{The camera orbits, but the unseen side is flattened, or the geometry splits in two and merges again~(\req{2}, \req{3}).} \\
    GenXD & 0.00\% & 75.36$^\circ$ & 0.00\% & 0.00$^\circ$ & 8 \\
    \sees{The design morphs continuously~(\req{2}, \req{3}), the surroundings change around the orbit~(\req{1}), and the orbit is incomplete~(\req{4}).} \\
    \midrule
    Stable Virtual Camera & 25.00\% & 303.55$^\circ$ & 20.50\% & 309.81$^\circ$ & 1450 \\
    \sees{The geometry changes shape during the orbit~(\req{2}, \req{3}), or the camera does not orbit the subject~(\req{4}).} \\
    GEN3C & 0.50\% & 200.22$^\circ$ & 0.90\% & 179.45$^\circ$ & 818 \\
    \sees{The camera orbits widely and loses the subject~(\req{4}). The geometry changes shape and disappears~(\req{2}, \req{3}).} \\
    ReCamMaster & 0.00\% & 125.75$^\circ$ & 0.00\% & 107.64$^\circ$ & 743 \\
    \sees{The camera drifts off the subject and leaves the orbit incomplete~\req{4}, so the side and rear views are never shown.} \\
    ViewCrafter & 0.00\% & 123.52$^\circ$ & 0.50\% & 152.87$^\circ$ & 712 \\
    \sees{The camera barely moves~(\req{4}), and the shape and surroundings keep changing in place~(\req{1}, \req{2}, \req{3}).} \\
    TrajectoryCrafter & 0.00\% & 93.60$^\circ$ & 0.00\% & 122.82$^\circ$ & 437 \\
    \sees{The geometry is hallucinated~(\req{2}, \req{3}), the surroundings reshuffle~(\req{1}), and the orbit is incomplete~(\req{4}).} \\
    \midrule
    TRELLIS.2 & \multicolumn{4}{c}{by construction} & 1856 \\
    \sees{One design with some diversity, sometimes structurally incomplete~\req{2}, and no surroundings~\req{1}.} \\
    Hunyuan3D-2 & \multicolumn{4}{c}{by construction} & 1263 \\
    \sees{One design with some diversity, sometimes structurally incomplete~\req{2}, and no surroundings~\req{1}.} \\
    \midrule
    Wan2.1-FLF2V-14B & 0.00\% & 28.77$^\circ$ & 0.00\% & 52.98$^\circ$ & 72 \\
    \sees{The camera barely moves, so the side and rear views are never shown~(\req{2}, \req{4}).} \\
    Wan2.1-I2V-14B & 0.00\% & 42.11$^\circ$ & 0.00\% & 73.10$^\circ$ & 570 \\
    \sees{The camera pans and stops, or zooms in and out, so the side and rear views are never shown~(\req{2}, \req{4}).} \\
    HunyuanVideo-I2V-13B & 0.00\% & 59.38$^\circ$ & 0.00\% & 86.23$^\circ$ & 494 \\
    \sees{The camera pans and stops, zooms in and out, or cuts abruptly to a different view, so the side and rear views are never shown~(\req{2}, \req{3}, \req{4}).} \\
    \bottomrule
  \end{tabular*}
  \label{tab:cross_model_compare2}
\end{table*}

\subsection{Novel View Synthesis with Gaussian Splatting}
\label{sec:novel_view_synth}

We further assess geometric consistency~\req{3} by reconstructing a 3D scene from each generated video and evaluating its ability to reproduce held-out views.
We attempt reconstruction for approximately 70 generated architectural videos per method. Videos for which COLMAP
reconstruction fails are excluded from this 3DGS evaluation, and metrics are averaged over the successfully reconstructed cases. For each remaining video, we fit 3D Gaussian Splatting
(3DGS)~\citep{kerbl3Dgaussians} to the raw orbit video, without additional video post-processing.

Half of the frames are used for fitting, while the remaining frames
are held out for evaluation. We render the reconstructed scene at
the held-out viewpoints and compare the renderings with the
corresponding generated frames using PSNR, SSIM, and LPIPS.

As shown in Table~\ref{tab:gs}, \method{} achieves the highest PSNR and SSIM and the lowest LPIPS among the evaluated methods. 
These results suggest that our generated views can be reproduced more faithfully by a single fitted 3D representation.
Since the reference images are held-out generated frames, this evaluation provides evidence of geometric consistency~\req{3}. 
DimensionX ranks second despite flattening unseen sides, so this evaluation complements the shape-consistency ratings~(\req{3}) of the user study (main-paper Tab.~\ref{tab:user_study}).

\begin{table}
  \centering
  \caption{\textbf{Novel view synthesis with 3D Gaussian Splatting.}
  For about 70 architecture inputs, we fit 3DGS~\citep{kerbl3Dgaussians} to each method's raw orbit video (no post-processing), using half of the frames for fitting and the other half for evaluation.
  \method{} scores best on all three metrics.}
  {\fontsize{8}{9.5}\selectfont
  \begin{tabular*}{\columnwidth}{@{\extracolsep{\fill}}lccc@{}}
    \toprule
    Method & PSNR $\uparrow$ & SSIM $\uparrow$ & LPIPS $\downarrow$\\
    \midrule
    \textbf{DesignOrbit}  &
    \textbf{21.49} &
    \textbf{0.74} &
    \textbf{0.17} \\

    DimensionX &
    20.46 &
    0.69 &
    0.20 \\

    Stable Virtual Camera &
    17.24 &
    0.57 &
    0.28 \\
    
    GEN3C &
    15.80 &
    0.48 &
    0.36 \\

    \bottomrule
  \end{tabular*}
  }
\label{tab:gs}
\end{table}

\begin{table}[t]
  \centering
  \small
  \setlength{\tabcolsep}{3pt}
  \caption{\textbf{Video generation runtime on one NVIDIA H200 NVL.} \method{} requires approximately 28\% less time than Stable Virtual Camera and 27\% less than GEN3C. DimensionX is faster in the evaluated configuration, but uses a lower spatial resolution and has lower full-orbit rates (Table~\ref{tab:cross_model_compare2}). Steps are per sampling call; time includes preprocessing, conditioning, all sampling calls, and decoding.}
  \label{tab:runtime}
  \begin{tabular*}{\linewidth}{@{\extracolsep{\fill}}lcrrr@{}}
    \toprule
    Method & Resolution & Frames & Steps & Time (s) $\downarrow$ \\
    \midrule
    DesignOrbit           & $832\times480$  &  81 & 30 & 242.3 \\
    DimensionX            & $480\times320$  & 145 & 50 & 182.4 \\
    Stable Virtual Camera & $1024\times576$ &  81 & 50 & 337.6 \\
    GEN3C                 & $1280\times704$ & 121 & 35 & 331.6 \\
    \bottomrule
  \end{tabular*}
\end{table}

\subsection{Runtime}
\label{sec:runtime}

Generation time matters at the concept stage, where proposals are revised continuously (main-paper \cref{sec:intro}). We measure end-to-end
generation time on a single NVIDIA H200 NVL, including preprocessing, conditioning, all sampling calls, and decoding.
Table~\ref{tab:runtime} reports the runtime and generation settings for each method.

\method{} generates an 81-frame video at $832\times480$ in
$242.3$ seconds, requiring approximately $28\%$ less time than
Stable Virtual Camera and $27\%$ less than GEN3C.
DimensionX is faster at $182.4$ seconds, producing 145 frames at
the lower spatial resolution of $480\times320$, but 
is less preferred on plausibility~\req{2} and consistency~\req{3} (main-paper Tab.~\ref{tab:user_study})
and completes only $6.3\%$ of architectural orbits~\req{4}.
These timings reflect the evaluated configurations; differences
in resolution, frame count, and sampling steps should be considered
when comparing computational efficiency.

\section{Implementation Details}

\subsection{Synthetic Orbit Data}
\label{sec:syn_data}

To train \method{} to generate $360^\circ$ orbital videos from a single concept image $I_0$, we construct synthetic videos
$\{I_0,\ldots,I_{T-1}\}$ whose cameras complete a full revolution and return to their starting viewpoint, with $I_{T-1}=I_0$.

\begin{figure*}[t]
\includegraphics[width=\linewidth]{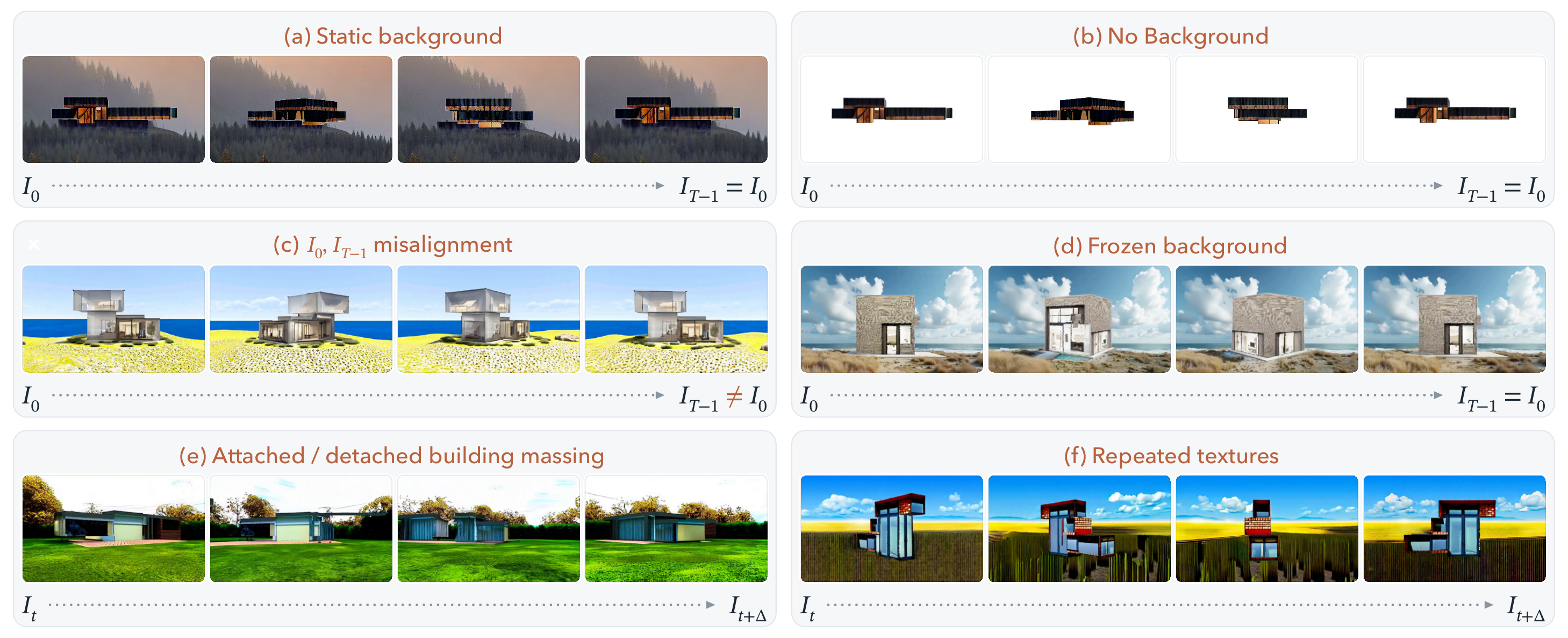}
\caption{\textbf{Synthetic-data construction and failure modes.}
Procedural building renders~\cite{Tsai20253D} with a static background (a), or with the background removed (b), do not provide training examples of camera-induced motion in the surroundings.
Single-pass background inpainting~\cite{vace} can yield mismatched endpoint images (c), while conditioning on identical first and last frames can freeze the background (d).
Split-and-condition inpainting (main-paper Fig.~\ref{fig:method}) combines a
text-conditioned first half with a second half anchored at the
completed midpoint and first frame. The endpoints match, but intermediate frames may still show unwanted structures (e) and repeated textures (f).
We select synthetic training clips using the verifier $V$
(Sec.~\ref{sec:camera_coverage}), which checks full-frame registration, camera coverage, directional agreement, and radius consistency.}
\label{fig:sync_data_problem}
\end{figure*}

Following~\cite{Tsai20253D}, we procedurally model cuboid buildings in 3D and render orbit trajectories with textured facades. 
The rendered building views are geometrically consistent~\req{3} and show architecturally plausible side and rear views~\req{2} by construction, and the camera returns to its starting pose~\req{4}.
However, renders following~\cite{Tsai20253D} have a static background, and removing it leaves the building with no background (Fig.~\ref{fig:sync_data_problem}a, b). Training on either version produces a model that rotates only the building while nothing behind it moves, so the site shows no parallax and gives no usable context~\req{1}.

To add camera-induced background motion while preserving the rendered
building, 
we mask everything except the rendered building and use a video diffusion inpainting model~\cite{vace} to synthesize surroundings.
The rendered foreground anchors the building geometry~\req{3}, while the inpainting model supplies the scene context~\req{1}.

\paragraph{The loop closure challenge.}
Inpainting all $T$ frames in a single pass can produce temporally smooth backgrounds without matching the first and last views~\req{4}. Conversely, conditioning the entire sequence on identical endpoint images can produce a frozen background with little parallax~\req{1}.

\paragraph{Split-and-condition inpainting.}
We split the orbit at $m=\lfloor(T-1)/2\rfloor$ and inpaint the two
halves with different boundary conditions (main-paper Fig.~\ref{fig:method}).

\noindent\textit{First half} ($I_0,\ldots,I_m$). We inpaint the masked background using a text prompt describing the surroundings, while preserving the rendered building. Without a repeated endpoint constraint, this pass can generate camera-induced background motion.

\noindent\textit{Second half} ($I_m,\ldots,I_{T-1}$).
We use the completed midpoint frame as the starting anchor and the
completed first frame as the ending anchor. We set $I_{T-1}=I_0$
and protect both boundary frames with masks $M'_m$ and $M'_{T-1}$,
while also preserving the rendered building in the intermediate frames. This constrains the second half to join the midpoint context
and return to the first view.

The two-pass construction enforces matching endpoint images while allowing background motion within the orbit. 
Intermediate frames can still contain attached or detached building structures and other inconsistent geometry~(\req{3}), or unrealistic surroundings~(\req{1}), motivating verification of the completed clips.

\paragraph{Failure modes and verification.} Figure~\ref{fig:sync_data_problem} illustrates the construction variants and their failure modes. Static backgrounds and background removal fail to provide camera-induced motion in the surroundings (a, b; \req{1}); single-pass inpainting can break loop closure (c; \req{4}), while direct conditioning on identical endpoints can freeze the background (d; \req{1}). Split-and-condition inpainting addresses these endpoint and motion problems but can still introduce unwanted structures (e; \req{3}) or repeated textures (f; \req{1}). We therefore apply the SfM-based verifier $V$ (main-paper \cref{sec:verifier}; implementation details in Sec.~\ref{sec:camera_coverage}) to select clips with complete frame registration, sufficient coverage, and consistent camera trajectories. These checks provide evidence of geometric consistency~\req{3} and orbit completeness~\req{4}.

\paragraph{Accepted synthetic training clips.}
Let $\mathcal X_{\mathrm{syn}}$ denote the candidate synthetic
composites after split-and-condition inpainting. The accepted
synthetic training set is
\[
\mathcal D_{\mathrm{syn}}
=\{x\in\mathcal X_{\mathrm{syn}}:V(x)=1\}.
\]
This set contains approximately 300 clips used to train Stage~I and
reused in Stage~II. Section~\ref{sec:sampling} reports the generated and retained counts.

\subsection{Sampling budgets and acceptance rates}
\label{sec:sampling}

For Stages~I and~II, we report the number of newly generated candidate
videos and the number retained by the training-data verifier $V$
(Sec.~\ref{sec:camera_coverage}). The acceptance rate is the fraction
of newly generated candidates that pass $V$; clips reused from an
earlier stage are not included in this denominator.

\emph{Stage~I:} We generate $\sim$2160 synthetic composites and retain
$\sim$300 as $\mathcal{D}_\text{syn}$; 13.9\% pass $V$. Overall, 74.3\%
register all 81 frames. Among these fully registered videos, 83.7\% pass
the radius-consistency check, 92.4\% pass the directional-agreement check,
and 18.7\% satisfy all verification criteria, including $C(x)\geq351^\circ$. None of the approximately 300 videos sampled from
the base model passes $V$.

\emph{Stage~II:} We generate $\sim$3125 videos from the Stage~I model on free-form building inputs and retain $\sim$300; 9.6\% pass $V$. Overall, 69.7\% register all 81 frames. Among these fully registered videos, 95.4\% pass the radius-consistency check, 100.0\% pass the
directional-agreement check, and 13.8\% satisfy all verification
criteria, including $C(x)\geq351^\circ$. We combine these newly verified clips with the $\sim$300 synthetic clips in $\mathcal{D}_\text{syn}$,
yielding $\sim$600 training videos, and train a fresh LoRA from the base
checkpoint.

\emph{Stage~III:}
For DPO, we generate $K=8$ videos for each of 360 image--text inputs,
yielding 2,880 candidate videos, 90\% of which register all 81 frames. We retain 350 winner--loser pairs
with a coverage margin of at least $\delta=30^\circ$ and split them
by input into 297 training pairs and 53 validation pairs.
Unlike Stages~I and~II, this stage selects preference pairs by relative
coverage rather than requiring each video to pass $V$.

\subsection{Hyperparameters} 
\label{sec:hyperparameter}

We fine-tune Wan2.1-FLF2V-14B~\cite{wan2025} at $832{\times}480$ and 81 frames, conditioning on the input image as both the first and the last frame ($I_0 = I_{T-1}$) and on a text prompt describing the scene. All LoRA~\cite{hu2022lora} adapters are applied to the attention projections and feed-forward layers of the DiT. The SFT stages use rank 32 and learning rate $1{\times}10^{-4}$; we keep the checkpoint that maximizes pass rate $\times$ mean coverage on held-out inputs. For DPO, we merge the Stage~II LoRA into the base weights, add a zero-initialized rank-16 LoRA, and train with $\beta{=}100$, $\lambda{=}0$, and learning rate $4{\times}10^{-5}$, accumulating gradients over 8 preference pairs per optimizer step. We draw timesteps only from the first half of the training schedule, which at Wan's scheduler shift of 3 corresponds to noise levels $t\in[0.75,1]$; global structure and camera motion are determined in this range, which govern completeness~\req{4}.

HunyuanVideo-I2V-13B follows the same recipe at $608{\times}352$ with last-frame conditioning added. Its scheduler shift is 7, so we draw from the first 70\% of the schedule to cover the same $t\in[0.75,1]$. We keep its DPO LoRA in fp32, because bf16 rounds away the small updates.

\subsection{Camera Coverage Computation} 
\label{sec:camera_coverage}

Coverage $C(x)$ measures orbit completeness~\req{4}.
For each video $x$, we recover camera poses from the registered images using COLMAP~\cite{schoenberger2016sfm,schoenberger2016mvs}.
Let $N_{\mathrm{reg}}(x)$ be the number of registered frames and
$T=81$ the total number of frames. Each pose consists of a rotation
$R_i$ and translation $t_i$ defining the world-to-camera mapping
$X_{\mathrm{camera}}=R_iX_{\mathrm{world}}+t_i$. The corresponding camera center is
\[
\mathbf c_i=-R_i^\top t_i.
\]
We compute coverage from these centers as follows.

\begin{enumerate}
\item \textbf{Projection and orbit center.}
We fit a plane to the mean-centered camera positions using singular
value decomposition (SVD) and project the centers onto that plane,
obtaining 2D positions $(x_i,y_i)$. We initialize the orbit center
$(c_x,c_y)$ at their centroid and refine it with one least-squares
step that reduces the spread of camera-to-center distances.

\item \textbf{Angles and camera filtering.}
We compute camera azimuths in degrees,
\[
\theta_i=\operatorname{atan2}(y_i-c_y,x_i-c_x)
\bmod 360^\circ,
\]
and angular steps in temporal order, accounting for wraparound.
For coverage computation, we retain cameras associated with steps
that agree in sign with the median step and are not outliers in step
size, and whose camera-to-center distances are not outliers.
Step-size and radius outliers are identified using three median
absolute deviations. Let $N_{\mathrm{keep}}(x)$ denote the number of
cameras remaining after this filtering. If fewer than eight remain,
we set $C(x)=0^\circ$.

\item \textbf{Gap measurement.}
For $N=N_{\mathrm{keep}}(x)\geq8$, we sort the retained azimuths as
$\theta_{(1)}\leq\cdots\leq\theta_{(N)}$ and compute
\[
\begin{aligned}
\Delta_i&=\theta_{(i+1)}-\theta_{(i)}, &&1\leq i<N,\\
\Delta_N&=\theta_{(1)}+360^\circ-\theta_{(N)}.
\end{aligned}
\]

\item \textbf{Coverage.}
The largest gap is the largest azimuth interval between retained
camera positions. We define
\[
C(x)=360^\circ-\max_i\Delta_i.
\]
\end{enumerate}

Higher coverage indicates more complete sampling around the estimated orbit center. For an ideal closed orbit with $T=81$ frames, the final viewpoint repeats the first, leaving $T-1=80$ distinct, uniformly spaced viewpoints. Thus,
\[
\Delta_{\mathrm{ideal}}=\frac{360^\circ}{T-1}=4.5^\circ,
\qquad
C_{\mathrm{ideal}}=355.5^\circ.
\]

The threshold $C(x)\geq351^\circ$ permits a largest azimuth gap of at most $9^\circ$, twice the ideal spacing, so at most one viewpoint is missing between any two neighboring retained cameras; this is the single missed frame of main-paper \cref{sec:verifier}. The bound applies to the largest gap, not to the total number of missing viewpoints.
The value $355.5^\circ$ describes the ideal repeated-endpoint sampling
pattern.

\paragraph{Training-data verifier.}
Main-paper \cref{sec:verifier} states the two primary checks (full registration and coverage); the directional-agreement and radius-consistency checks 
complete the verifier used in Stages~I and~II.
Full registration checks consistency~\req{3}; coverage, directional agreement, and radius consistency together check completeness~\req{4}: the
camera goes around the full circle, in one direction, at a near-constant distance.
Let $A(x)$ be the fraction of temporal angular steps that have the
same sign as the median step, and let
$\mathrm{CV}_r(x)=\sigma_r/\mu_r$ be the coefficient of variation of
the camera-to-center distances. Both statistics are computed before
camera filtering: $A(x)$ uses steps between temporally consecutive
registered cameras, and $\mathrm{CV}_r(x)$ uses all registered camera radii.

A video is accepted, $V(x)=1$, if and only if $N_{\mathrm{reg}}(x)=T=81, C(x)\geq351^\circ, A(x)\geq0.90, \mathrm{CV}_r(x)\leq0.30$. Otherwise, $V(x)=0$. 

The binary verifier is used to select training videos for Stages~I
and~II. In the final DPO configuration, videos are instead ranked by $S(x)=C(x)$ when coverage is computable, with $S(x)=0^\circ$ otherwise; constructing a preference pair does not require either video to satisfy $V(x)=1$.

\end{document}